\documentclass[lettersize,journal]{IEEEtran}
\usepackage{amsmath,amsfonts}
\usepackage{algorithmic}
\usepackage{algorithm}
\usepackage{array}
\usepackage[caption=false,font=normalsize,labelfont=sf,textfont=sf]{subfig}
\usepackage{textcomp}
\usepackage{stfloats}
\usepackage{url}
\usepackage{verbatim}
\usepackage{graphicx}
\usepackage{cite}
\usepackage{tabularx}
\usepackage{adjustbox}
\usepackage{booktabs}
\usepackage{multirow}
\usepackage{amsmath}
\usepackage{amssymb}
\usepackage{cuted}
\usepackage{comment}
\usepackage{lipsum}
\usepackage{color}
\usepackage{CJKutf8}
\usepackage{relsize}

\begin{document}

\title{EMS: Efficient and Effective Massively Multilingual Sentence Embedding Learning}

\author{Zhuoyuan~Mao,
        Chenhui~Chu,
        and~Sadao~Kurohashi
\thanks{The authors are with the Kyoto University, Kyoto 606-8501, Japan (e-mail:
zhuoyuanmao@nlp.ist.i.kyoto-u.ac.jp; chu@i.kyoto-u.ac.jp; kuro@i.kyoto-u.ac.jp).}
\thanks{Manuscript received *** **, 2023; revised *** **, 2023.}}

\markboth{Journal of \LaTeX\ Class Files,~Vol.~14, No.~8, August~2021}%
{Shell \MakeLowercase{\textit{et al.}}: A Sample Article Using IEEEtran.cls for IEEE Journals}

\IEEEpubid{0000--0000/00\$00.00~\copyright~2021 IEEE}

\maketitle

\begin{abstract}
Massively multilingual sentence representation models, e.g., LASER, SBERT-distill, and LaBSE, help significantly improve cross-lingual downstream tasks. However, the use of a large amount of data or inefficient model architectures results in heavy computation to train a new model according to our preferred languages and domains. To resolve this issue, we introduce efficient and effective massively multilingual sentence embedding (EMS), using cross-lingual token-level reconstruction (XTR) and sentence-level contrastive learning as training objectives. Compared with related studies, the proposed model can be efficiently trained using significantly fewer parallel sentences and GPU computation resources. Empirical results showed that the proposed model significantly yields better or comparable results with regard to cross-lingual sentence retrieval, zero-shot cross-lingual genre classification, and sentiment classification. Ablative analyses demonstrated the efficiency and effectiveness of each component of the proposed model. We release the codes for model training and the EMS pre-trained sentence embedding model, which supports 62 languages (\url{https://github.com/Mao-KU/EMS}).


\end{abstract}



\section{Introduction}\label{sec:introduction}
\IEEEPARstart{C}{ROSS-LINGUAL} sentence representation (CSR) models~\cite{DBLP:conf/rep4nlp/SchwenkD17,DBLP:journals/jstsp/Espana-BonetVBG17,DBLP:conf/rep4nlp/YuLO18,DBLP:conf/naacl/DevlinCLT19,DBLP:conf/rep4nlp/ChidambaramYCYS19,DBLP:journals/tacl/ArtetxeS19,DBLP:journals/corr/abs-1912-12481,DBLP:conf/nips/ConneauL19,DBLP:conf/ijcai/YangAYGSCSSK19,DBLP:conf/acl/YangCAGLCAYTSSK20,DBLP:conf/emnlp/ReimersG20,DBLP:conf/acl/ConneauKGCWGGOZ20,DBLP:conf/acl/MaoGCJK20,DBLP:conf/acl/FengYCA022,DBLP:conf/eacl/MaoN23,DBLP:conf/eacl/ZhaoWCT24} prove to be essential for downstream NLP tasks like cross-lingual sentence retrieval and cross-lingual transfer without the need for initial training and monolingual model. Thus, CSR models benefit low-resource languages without sufficient training data.

A majority of the CSR training methods can be ascribed to one of the following two categories: \textit{global fine-tuning} or \textit{sentence embedding}. \textit{global fine-tuning} methods indicate that for a specific downstream task, we conduct fine-tuning by updating pre-trained language models e.g., mBERT~\cite{DBLP:conf/naacl/DevlinCLT19}, XLM~\cite{DBLP:conf/nips/ConneauL19}, and XLM-R~\cite{DBLP:conf/acl/ConneauKGCWGGOZ20}. The fine-tuning efficiency of this method group is determined by the scale of the pre-trained model. Thus, the update of the large-scale parameters of the pre-trained model tends to be the computation bottleneck for fine-tuning. The computationally lite \textit{global fine-tuning} methods have been explored sufficiently either by compressing the model~\cite{DBLP:conf/iclr/LanCGGSS20}, training a student by knowledge distillation~\cite{DBLP:journals/corr/abs-1910-01108,DBLP:conf/acl/SunYSLYZ20,DBLP:conf/emnlp/JiaoYSJCL0L20}. On the other hand, \textit{sentence embedding} methods, e.g., LASER~\cite{DBLP:journals/tacl/ArtetxeS19}, aim to train the CSR that aligns the embedding space across languages without further fine-tuning.
For example, the English sentence ``\underline{I am a student.}'' should have an identical sentence embedding to its French translation, ``\underline{Je suis un étudiant.}''
As a result, this group of methods can be efficiently adapted to several cross-lingual downstream tasks by merely adding a multi-layer perceptron without the need for tuning parameters within the pre-trained CSR model. Sentence embedding is a vital technique in practical applications. For example, it enables the extraction of parallel data from extensive web corpora across two languages, enhancing the training of high-quality machine translation and multilingual language models~\cite{DBLP:conf/eacl/SchwenkCSGG21}. Furthermore, sentence embedding is instrumental for zero-shot cross-lingual retrieval, an essential method for enabling cross-lingual searches on e-commerce platforms like Amazon~\cite{DBLP:conf/www/XuY23}. However, existing \textit{massively multilingual sentence embedding (MSE)} models, LASER~\cite{DBLP:journals/tacl/ArtetxeS19}, SBERT-distill~\cite{DBLP:conf/emnlp/ReimersG20}, and LaBSE~\cite{DBLP:conf/acl/FengYCA022}, require a considerable amount of data or inefficient model architectures, for which the efficient training objectives have not been explored.

\IEEEpubidadjcol

In this study, we present \textbf{E}fficient and effective massively \textbf{M}ultilingual \textbf{S}entence embedding (\textbf{EMS}), a computationally lite and effective architecture for training MSE, which ameliorates the data and computation efficiency to train an MSE model according to our preferred domains or language groups and may have a promising future for deploying MSE model training and adaptation on memory-limited devices. In particular, we propose cross-lingual token-level reconstruction (XTR) and sentence-level contrastive learning as training objectives. XTR captures the target token distribution information, whereas the contrastive objective serves to recognize translation pairs. We claim that these two objectives effectively construct the multilingual signals for learning MSE within the dual-encoder model architecture, which results in highly efficient model training. Compared with previous MSE models in the massively multilingual scenario, EMS can be trained using significantly fewer parallel data and less GPU consumption.

In contrast to our previous study~\cite{DBLP:conf/acl/MaoGCJK20}, lightweight bilingual sentence representation learning, we focus on exploring how to train a model efficiently and effectively for a massively multilingual scenario in this work. To address this, we tailor the model capacity for a large number of languages and introduce a language embedding layer for the generative objective and MLP layers for the contrastive objective. Furthermore, our findings indicate that the combination of the XTR objective and the alignment-based sentence-level contrastive objective, as proposed in our previous study, is advantageous for massively multilingual training. In contrast, the unified generative task (UGT) from our earlier work does not perform effectively in such a scenario. Notably, we discovered that the sentence-level contrastive objective consistently enhances performance in cross-lingual retrieval and classification tasks as jointly trained with XTR. This contrasts with our previous observation, where this objective was detrimental to classification tasks in bilingual sentence embedding models. In addition, regarding model performance, we validate the effectiveness of EMS with over 100 languages and more evaluation benchmarks in this study, along with the contribution of each model component via the ablation study.

Despite the small amount of training data and low-cost training, experimental results demonstrate that the proposed EMS learned a robustly aligned multilingual sentence embedding space. With regard to the Tatoeba~\cite{DBLP:journals/tacl/ArtetxeS19} cross-lingual similarity benchmark, EMS significantly achieves better results than LASER and SBERT-distill and comparable results considering middle- and high-resource languages\footnote{languages for which we possess over 300k parallel sentences for training data.} compared with LaBSE. Based on the results on Flores~\cite{DBLP:journals/tacl/GoyalGCCWJKRGF22} cross-lingual similarity benchmark for non-English language pairs, we demonstrate that EMS is completely language-agnostic while LASER is an English-dependent model. Moreover, we evaluate the model performance for mining parallel sentences from larger comparable corpora, including the task of ParaCrawl~\cite{DBLP:conf/acl/BanonCHHHEFKKKO20} sentence retrieval and BUCC benchmark~\cite{DBLP:conf/acl-bucc/ZweigenbaumSR17,zweigenbaum2018overview}. The experimental results show that EMS performs better than SBERT-distill and comparably with LASER. Furthermore, we evaluate the language-agnostic representation based on three classification tasks in a zero-shot manner, document genre classification based on MLDoc~\cite{DBLP:conf/lrec/SchwenkL18}, and sentiment classification based on two Amazon review datasets~\cite{DBLP:conf/acl/PrettenhoferS10,DBLP:conf/emnlp/KeungLSS20}. Empirical results show that EMS outperforms LASER and SBERT-distill on MLDoc and one of the Amazon review datasets and yields comparable performance with SBERT-distill and LaBSE on the other Amazon review dataset. In addition, upon integrating LaBSE's additive margin softmax (AMS) contrastive objective into the EMS framework, while maintaining identical training data and model architecture, we noted a decline in performance. This outcome suggests that the effectiveness of AMS's objective is heavily reliant on LaBSE's extensive batch size and training data. It also highlights the superior efficacy of the form of the contrastive objective proposed in our study and the complementary nature of the XTR generative objective.


The major contributions of this paper are summarized as:
\begin{itemize}
    \item The training architecture and objectives we developed were both efficient in terms of data and computation, and they achieved improved or competitive results in cross-lingual sentence retrieval and sentence classification tasks when compared to other MSE models.
    \item We identified effective forms of generative and contrastive objectives, and demonstrated that the proposed language embedding layers significantly enhance MSE performance in massively multilingual scenarios, marking a notable advancement from our previous study in bilingual settings.
    \item We revealed that incorporating MLP layers prior to loss computation and combining with generative objectives can enhance LaBSE's AMS loss for MSE learning.
    \item We release the codes of the model training and the EMS model, which supports 62 languages.
\end{itemize}

\section{Related Work}
\label{sec:2}

In this section, we revisit the literature on recent MSE models and training objectives for developing MSE.


\subsection{Multilingual Sentence Embedding}
\label{sec:2.1}
The pursuit of dense text embeddings has evolved significantly, beginning with the advent of word vectors~\cite{DBLP:journals/corr/abs-1301-3781} and progressing to sentence embeddings. Initial approaches, such as those by~\cite{DBLP:conf/iclr/AroraLM17}, advocated for the weighted average of word embeddings to create sentence embeddings, establishing a robust baseline. Subsequent efforts shifted towards leveraging neural models~\cite{DBLP:conf/emnlp/ConneauKSBB17,DBLP:conf/emnlp/CerYKHLJCGYTSK18} and pre-trained transformers~\cite{DBLP:conf/emnlp/ReimersG19,DBLP:conf/acl/NiACMHCY22,DBLP:conf/emnlp/YangYCLD21,DBLP:journals/corr/abs-2212-03533} as backbone architectures. Recent studies have predominantly focused on refining the training objectives with contrastive loss~\cite{DBLP:conf/emnlp/ZhangHLLB20,DBLP:conf/acl/GiorgiNWB20,DBLP:conf/acl/KimYL20,DBLP:conf/acl/YanLWZWX20,DBLP:conf/emnlp/GaoYC21,DBLP:conf/acl/ChengYSLQ23} and on the strategic use of various training datasets, often involving translation pairs~\cite{DBLP:conf/acl/0004HLB020}. More recently, the integration of large language models (LLMs)~\cite{DBLP:journals/corr/abs-2307-16645} and prompting methods~\cite{DBLP:conf/acl/SuSKWHOYSZ023} have further advanced the capabilities of sentence embedding models.

In the realm of multilingual contexts,~\cite{DBLP:conf/rep4nlp/SchwenkD17} pioneered the concept of MSE, leveraging intermediate representations from LSTM~\cite{DBLP:journals/neco/HochreiterS97} encoder-decoder frameworks in neural machine translation (NMT). Concurrently,~\cite{DBLP:journals/corr/abs-1709-09783} devised MSE by aligning outputs from LSTM dual encoders (akin to Siamese networks~\cite{DBLP:journals/tacl/YinSXZ16}) into a unified representational space. Building on this,~\cite{DBLP:journals/jstsp/Espana-BonetVBG17} experimented with sum pooling of NMT encoder's top hidden states, diverging from the max pooling and last hidden state approach in~\cite{DBLP:conf/rep4nlp/SchwenkD17}.~\cite{DBLP:conf/rep4nlp/YuLO18} introduced a training methodology for MSE that combines bidirectional NMT losses and minimizes the Euclidean distance between translation pair embeddings.

The transition to dual transformer architectures replacing LSTM was initiated by~\cite{DBLP:conf/wmt/GuoSYGCASCSSK18}, who first utilized transformers for constructing MSE in bilingual settings. This was expanded by~\cite{DBLP:conf/rep4nlp/ChidambaramYCYS19}, who incorporated multiple tasks such as conversational response~\cite{DBLP:conf/rep4nlp/YangYCKCPGSSK18}, quick-thought~\cite{DBLP:conf/iclr/LogeswaranL18}, natural language inference~\cite{DBLP:conf/emnlp/BowmanAPM15}, and translation into the training regimen. Building on these efforts,~\cite{DBLP:conf/ijcai/YangAYGSCSSK19} further refined the training objectives by integrating an AMS loss, enhancing the approach proposed by~\cite{DBLP:conf/wmt/GuoSYGCASCSSK18}.

Subsequently, research shifted towards massively multilingual contexts, aiming to develop universal sentence embedding models supporting a large number of languages, usually at least over 10 languages. mUSE~\cite{DBLP:conf/acl/YangCAGLCAYTSSK20} pioneered this effort by training MSE for 16 languages, adopting the training methodology of~\cite{DBLP:conf/rep4nlp/ChidambaramYCYS19}. Concurrently, LASER~\cite{DBLP:journals/tacl/ArtetxeS19} utilized an LSTM framework to train MSE for 93 languages, expanding upon~\cite{DBLP:conf/rep4nlp/SchwenkD17}. Following this,~\cite{DBLP:conf/emnlp/ReimersG20} introduced SBERT-distill, leveraging parallel sentences to distill multilingual capabilities from pre-trained English sentence encoders for 50 languages. LaBSE~\cite{DBLP:conf/acl/FengYCA022} further advanced~\cite{DBLP:conf/ijcai/YangAYGSCSSK19} by extending from bilingual scenarios to 109 languages using a pre-trained masked language model.

Concurrently, the training objectives for MSE continued to evolve.~\cite{DBLP:conf/emnlp/GoswamiDAFM21} proposed an unsupervised multi-task learning approach for training MSE, eliminating the reliance on parallel sentences. Despite this innovation, their results still fell short of massively multilingual supervised models like LASER and LaBSE in cross-lingual sentence retrieval tasks. mSimCSE~\cite{DBLP:conf/emnlp/WangWN22} adapted the English monolingual SimCSE~\cite{DBLP:conf/emnlp/GaoYC21} to multilingual contexts, achieving performance comparable to LASER and marginally below LaBSE. LEALLA~\cite{DBLP:conf/eacl/MaoN23} introduced a method for distilling robust low-dimensional MSE from LaBSE using knowledge distillation. This technique could similarly be applied to distill efficient MSE from our EMS model. Recently, research has been delving into constructing MSE using LLMs through contrastive objectives~\cite{DBLP:journals/corr/abs-2307-16645,DBLP:conf/acl/SuSKWHOYSZ023}.

Recent research has also focused on incorporating word-level supervision in the training of MSE alongside traditional sentence-level contrastive objectives. Our previous work~\cite{DBLP:conf/acl/MaoGCJK20} introduced and validated the efficacy of a word-level XTR objective in bilingual settings, and this paper extends that approach to a massively multilingual setting by incorporating a language embedding layer tailored for multilingual scenarios. Concurrently,~\cite{DBLP:conf/acl/LiHZDLHJWDZ23} developed a method for training MSE across 36 languages, introducing a representation translation learning task that utilizes contextualized token representations from one language to reconstruct their counterparts in another language. This method resonates with our focus on utilizing cross-lingual token-level signals to enhance MSE. Given the simultaneous development of these methods, a comparative analysis with their approach is reserved for future research.

In this study, we continue to focus on the exploration of effective objectives for training MSE in massively multilingual contexts. We introduce the token-level XTR and sentence-level contrastive objectives, ensuring enhanced training efficiency and effectiveness on downstream tasks. The subsequent subsection will detail the discussion of the training objectives for sentence embedding models.

\subsection{Training Objectives for Sentence Embedding Learning}
This section provides an in-depth survey of two training objective types usually used for constructing sentence embedding, followed by a comprehensive discussion on the current state of research regarding these objectives within the context of MSE.

\noindent
\textbf{Generative Objectives}
measure a generation probability of the token prediction, via training a language model, which primarily contributes to the performance of downstream tasks. BERT's masked language model (MLM)~\cite{DBLP:conf/naacl/DevlinCLT19} and its variants~\cite{DBLP:conf/nips/ConneauL19,DBLP:conf/emnlp/RenWLZM19,DBLP:conf/acl/ConneauKGCWGGOZ20} focused on optimizing the encoder-side token generation probability. Sequence-to-sequence learning used the encoder--decoder framework to train either a translation task~\cite{DBLP:conf/rep4nlp/SchwenkD17,DBLP:journals/jstsp/Espana-BonetVBG17,DBLP:journals/tacl/ArtetxeS19} or a sentence reconstruction task~\cite{DBLP:conf/icml/SongTQLL19,DBLP:journals/jmlr/RaffelSRLNMZLL20,DBLP:conf/acl/LewisLGGMLSZ20} through optimizing the decoder-side token generation probability. Subsequently, sentence embedding could be constructed using the encoder-side output for both groups of generative objectives.

\noindent
\textbf{Contrastive Objectives}
aim to transform the representation space by adjusting the distance between the representations of tokens (or the sentences), which were initially used jointly with the generative objectives to improve sentence representation learning. Next sentence prediction (NSP) in BERT~\cite{DBLP:conf/naacl/DevlinCLT19}, token discrimination in ELECTRA~\cite{DBLP:conf/iclr/ClarkLLM20}, sentence discrimination in DeCLUTR~\cite{DBLP:conf/acl/GiorgiNWB20}, and hierarchical contrastive objective in HICTL~\cite{DBLP:conf/iclr/WeiW0XYL21} were the typical ones. Recent research, notably the SimCSE~\cite{DBLP:conf/emnlp/GaoYC21} study, has shown exceptional results by focusing solely on training with contrastive objectives.

Referring to the evolution of MSE training objectives discussed in Section~\ref{sec:2.1}, the contrastive objective has been widely adopted in MSE research, including our prior work~\cite{DBLP:conf/acl/MaoGCJK20}. Yet, in massively multilingual contexts, the optimal variant of the contrastive objective remains uncertain. This study evaluates the MLP-based contrastive objective inheriting our earlier work and contrasts it with the AMS loss used in LaBSE. While generative objectives have typically relied on translation tasks~\cite{DBLP:journals/tacl/ArtetxeS19}, these can be inefficient. Token-level generative tasks built upon a dual-encoder framework offer greater efficiency and have been less investigated. Therefore, we introduce the XTR objective into training for massive MSE, enhancing it with a suitable form of contrastive objective through joint training, extending our previous research in the bilingual domain~\cite{DBLP:conf/acl/MaoGCJK20} to a massively multilingual scenario. The only other concurrent study employing a token-level generative objective is~\cite{DBLP:conf/acl/LiHZDLHJWDZ23}, as mentioned in Section~\ref{sec:2.1}. Another efficient paradigm of the generative objective for MSE was by knowledge distillation introduced in SBERT-distill~\cite{DBLP:conf/emnlp/ReimersG20}, which we treat as a baseline for comparison in this study.


\section{Proposed Methods}
\label{sec:3}

\begin{figure*}[t]
    \centering
    \includegraphics[width=0.9\linewidth]{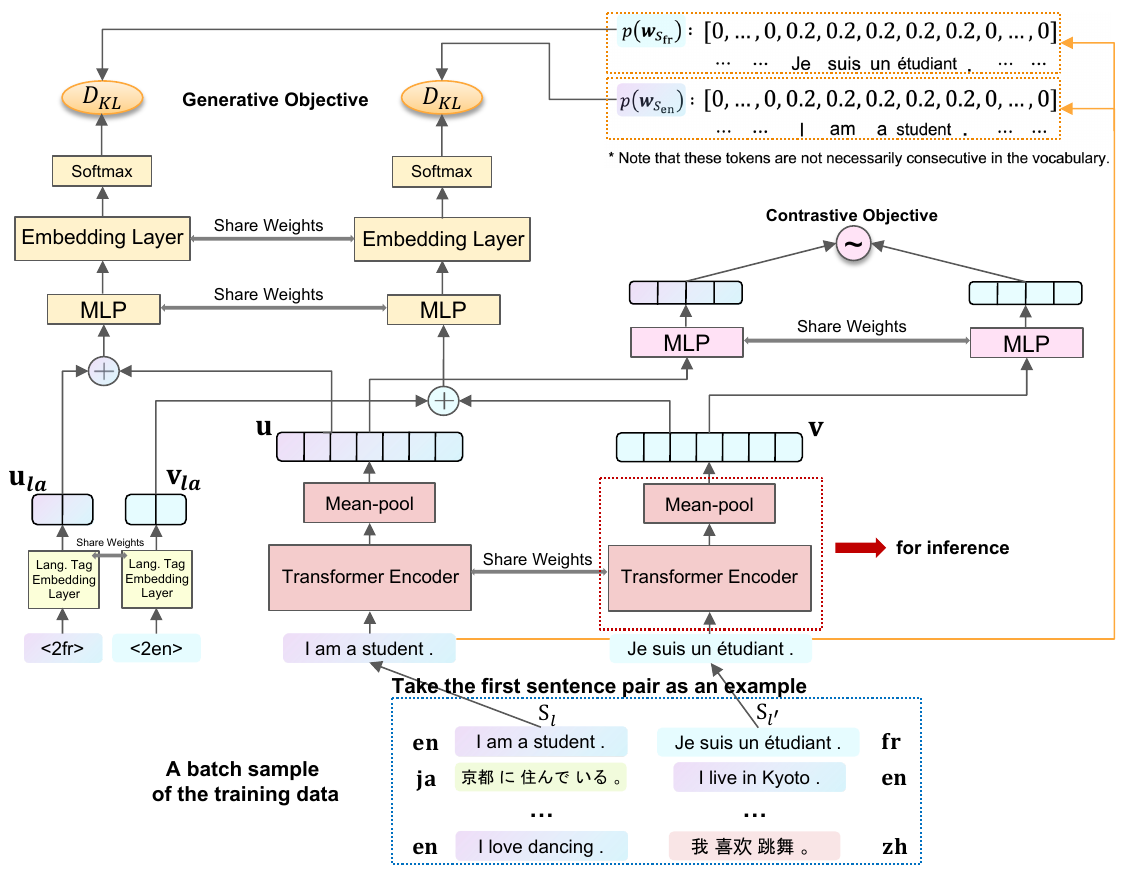}
    \caption{\textbf{Training architecture of EMS.} $\mathbf{u}$ and $\mathbf{v}$ are MSEs for inference, and the model components in the red dashed rectangle are used for inference. $\mathbf{u}_{la}$ and $\mathbf{v}_{la}$ are the target language embeddings. $\oplus$ denotes the hidden vector concatenation. The training data batch sample is given in the blue dashed box. Orange arrows and dashed box denote the gold token distributions within the generative objective, specifically the discrete uniform distributions for the tokens in $S_{en}$ and $S_{fr}$, denoted by $\mathlarger{p}_{S_{en}}$ and $\mathlarger{p}_{S_{fr}}$, respectively. The part within the red dashed box indicates the pre-trained EMS model for downstream tasks.}
    \label{architecture}
\end{figure*}


We conduct massively MSE learning by employing the dual transformer encoder as the backbone of the training framework. For the training objective, we propose a novel cross-lingual training method, which jointly optimizes generative and contrastive objectives. We introduce cross-lingual token-level reconstruction (XTR) as the generative objective and employ sentence-level self-supervised learning as the contrastive objective. The training framework and objectives that we propose are expected to learn a well-aligned representation space for multiple languages.

\subsection{Architecture}
\label{sec:3.1}

We introduce the dual transformer sharing parameters to encode parallel sentences along with several multi-layer perceptrons (MLP) to extract cross-lingual information and compute the generative and contrastive losses (Fig.~\ref{architecture}). We use parallel corpora as the training data. First, we build monolingual sentence representations $\mathbf{u}$ and $\mathbf{v}$ on top of a transformer encoder. Two groups of the MLP are employed to construct two training objectives. After completing the model training, given a sentence in any language, we use the transformer encoder to infer the language-agnostic sentence representation. We can implement cross-lingual downstream tasks in a zero-shot manner using $\mathbf{u}$ or $\mathbf{v}$, as they are representations independent of the specific language.

Specifically, as shown in Fig.~\ref{architecture}, assume that we have a parallel corpus $\mathbf{C}$ that includes multiple languages $\{l_1,l_2,...,l_N\}$, and each sentence pair $S=(S_{l}, S_{l^{\prime}})$ contains a sentence in language $l$ and its translation in language $l^{\prime}$, where $l, l^{\prime} \in \{l_1,l_2,...,l_N\}$, as shown in the blue dashed box in Fig.~\ref{architecture}. We use the dual transformer encoder $E$ sharing parameters to encode each sentence pair. Assume that the transformer encoder outputs of $S_{l}$ are $(\mathbf{h}^\mathrm{T}_{1},\mathbf{h}^\mathrm{T}_{2},...,\mathbf{h}^\mathrm{T}_{\parallel S_{l}\parallel})$, where $\parallel$$S_{l}$$\parallel$ indicates the length of $S_{l}$. We use the mean-pooled hidden states as the language-agnostic sentence representation $\mathbf{u}$:
\begin{equation}
    \mathbf{u} = \frac{1}{\parallel S_{l}\parallel}\sum_{i}{\mathbf{h}_{i}}
\end{equation}
Similarly, we can obtain $\mathbf{v}$ for $S_{l^{\prime}}$.


\subsection{Generative Objective}
\label{sec:3.2}
Generative objective plays an essential role in MSE learning. SBERT-distill and LaBSE use the pre-trained models as the model initialization; therefore, the pre-trained language models for each language serve as generative objectives. LASER finished the model training in one run without using any pre-training models, and the translation objective serves as a cross-lingual generative objective. Inspired by LASER, we include the generative objective for the one-run
model training. However, the presence of the transformer decoder in LASER increases the computational overhead. Instead, we propose a novel generative objective known as cross-lingual token-level reconstruction (XTR) to improve the training efficiency while retaining the quality of sentence representation, which circumvents using the transformer decoder.\footnote{It should be noted that both LASER and our model have the potential for further enhancement through language model pre-training. However, this aspect falls outside the scope of the current study and is left for future work.}

As we expect the XTR objective to measure a cross-lingual reconstruction loss, it is necessary to notify the model what the target language is. Thus, we compute a target language representation for each sentence by employing a language embedding layer $L_{la}$ to encode the target language token (e.g., $<2en>$ if the target language is English). More precisely, for each sentence pair $S=(S_{l}, S_{l^{\prime}})$,  
\begin{align}
    \mathbf{u}_{la} &= \mathbf{W}_{la}{\mathbf{h}_{l^{\prime}}} \\
    \mathbf{v}_{la} &= \mathbf{W}_{la}{\mathbf{h}_{l}}
\end{align}
where $\mathbf{W}_{la}\in \mathbb{R}^{d_{la}\times d_{vcb}}$ denotes the parameters of $L_{la}$. $\mathbf{h}_{l}$ and $\mathbf{h}_{l^{\prime}}$ respectively denote the one-hot embedding of $<2l>$ and $<2l^{\prime}>$. $d_{la}$ and $d_{vcb}$ denote the dimension of the language embedding and the size of the vocabulary, respectively. The incorporation of language embeddings effectively clarifies the target of the XTR objective, particularly when transitioning from our previous bilingual-focused study to a massively multilingual scenario.

Subsequently, we concatenate the language representation with the sentence representation and use a fully connected layer $L_{fc}$ to transform the concatenated representation for extracting the cross-lingual information. Finally, we use another linear embedding layer $L_{emb}$ followed by Softmax to transform the representation to present two probability distributions, which are formulated as:
\begin{align}
    \mathlarger{q}_{S_{l}} &= \mathrm{softmax}(\mathbf{W}_{emb}\sigma_{xtr}(\mathbf{W}_{fc}(\mathbf{u}_{la} \oplus \mathbf{u}) + \mathbf{b}_{fc})) \\
    \mathlarger{q}_{S_{l^{\prime}}} &= \mathrm{softmax}(\mathbf{W}_{emb}\sigma_{xtr}(\mathbf{W}_{fc}(\mathbf{v}_{la} \oplus \mathbf{v}) + \mathbf{b}_{fc}))
\end{align}
where $\mathbf{W}_{emb} \in \mathbb{R}^{d_{vcb}\times (d_{la}+d)}$, $\mathbf{W}_{fc} \in \mathbb{R}^{(d_{la}+d)\times (d_{la}+d)}$, $\mathbf{b}_{fc} \in \mathbb{R}^{d_{la}+d}$, and $d$ indicates the dimension of $\mathbf{u}$ (or $\mathbf{v}$). $\sigma_{xtr}$ is the activation function in $L_{fc}$, for which we use $\mathrm{swish}$~\cite{DBLP:conf/iclr/RamachandranZL18}. $\oplus$ indicates the concatenation over the first dimension. In our previous study~\cite{DBLP:conf/acl/MaoGCJK20}, we employed the identical parameters for $\mathbf{W}_{emb}$ as that in the transformer encoder. We demonstrate in this work that use different parameters for $\mathbf{W}_{emb}$ would enhance further enhance the MSE in the massively multilingual scenario (see Section~\ref{sec:5.7}).

Assume that $\mathbf{B}_{i}$ is a batch sampled from the training corpus $\mathbf{C}$. Then, the training loss of the XTR objective for the $\mathbf{B}_{i}$ is formulated as follows:
\begin{align}
\resizebox{\linewidth}{!}{ 
$\mathcal{L}_{XTR}^{(i)} =
\displaystyle\sum_{\substack{S \in \mathbf{B}_{i}}}
\Bigl(
\mathcal{D}_{KL}\left(\mathlarger{p}_{S_{l^{\prime}}}\left(\mathbb{W}\right) \parallel \mathlarger{q}_{S_{l}}\right)
+
\mathcal{D}_{KL}\left(\mathlarger{p}_{S_{l}}\left(\mathbb{W}\right) \parallel \mathlarger{q}_{S_{l^{\prime}}}\right)
\Bigr)
$}
\label{eq:xtr}
\end{align}
where $\mathcal{D}_{KL}$ denotes KL-divergence and $\mathbb{W}$ indicates the vocabulary set. As illustrated in the orange dashed box in Fig.~\ref{architecture}, we use discrete uniform distribution for the tokens in $S_l$ to define $\mathlarger{p}_{S_{l}}$.
Specifically, for each $w \in \mathbb{W}$, $\mathlarger{p}_{S_{l}}\left(w\right)$ is defined as:
\begin{equation}
\mathlarger{p}_{S_{l}}\left(w\right)=\left\{
\begin{aligned}
\frac{N_{w}}{\left\|S_{l}\right\|}&,& \hspace{1em} w\in S_{l}\\
0&,& \hspace{1em} w\notin S_{l}
\end{aligned}
\right.
\label{kl}
\end{equation}
where $N_{w}$ indicates the number of words $w$ in sentence $S_{l}$, and $N_{w}$ is 1 in most cases. $\left\|S_{l}\right\|$ indicates the length of $S_{l}$. In other words, $\mathlarger{p}_{S_{l}}\left(\mathbb{W}\right)$ is approximately an average of one-hot embeddings of $S_{l}$'s tokens. Similarly, we can obtain the definition of $\mathlarger{p}_{S_{l^{\prime}}}\left(\mathbb{W}\right)$.

Herein, we use the KL-divergence to measure the similarity between the token distribution of the sentence in the target language and the model output of the sentence in the source language and vice versa, which helps align the language-agnostic representation space. In Section~\ref{sec:5.7}, we will demonstrate that this objective also possesses good alignment abilities for non-English language pairs, even when trained on English-centric data, thanks to the exposure to multiple languages during the training process.

Moreover, in our previous study~\cite{DBLP:conf/acl/MaoGCJK20}, we introduced another generative objective known as the unified generative task (UGT) that combines XTR and single-word MLM~\cite{DBLP:conf/acl/MaoGCJK20}. We will provide empirical results and analyses to show that this objective is not relevant in the massively multilingual scenario and current model architecture (see Section~\ref{sec:5.7}).

\subsection{Contrastive Objective}
\label{sec:3.3}

Based on our previous study~\cite{DBLP:conf/acl/MaoGCJK20}, we employ a sentence-level contrastive objective as an assisting objective to force the model to grasp similar information of sentences across languages. We demonstrate that the sentence-level contrastive objective is a beneficial model component to jointly assist the generative objective. In Section~\ref{sec:5.7}, we provide empirical shreds of evidence that this objective plays a beneficial role in the generative objective introduced in Section~\ref{sec:3.2}.

Specifically, we employ in-batch sentence-level contrastive learning by discriminating between positive and negative samples for each sentence. Given a sentence, its translation (paired sentence in another language) is deemed as a positive sample, whereas other sentences within the batch are used as negative samples. Unlike our previous study, we employ temperature-based scaling and add two fully-connected layers to decrease the dimension of the sentence representation to compute the contrastive objective, following~\cite{DBLP:conf/icml/ChenK0H20}. Assume that $\mathbf{B}_{i}$ is a batch sampled from the training corpus $\mathbf{C}$, and the j-th sentence pair of $\mathbf{B}_{i}$ is $S^{(ij)}=(S_{l}^{(ij)},S_{l^{\prime}}^{(ij)})$. Then the sentence-level contrastive objective for $\mathbf{B}_{i}$ is formulated as:
\begin{equation}
\begin{split}
\mathcal{L}_{cntrs}^{(i)} =
-\sum_{S^{(ij)}\in\mathbf{B}_{i}}\Bigl(\log\frac{\exp{(\mathrm{sim}(S_{l}^{(ij)},S_{l^{\prime}}^{(ij)})/T)}}{\displaystyle\sum_{S^{(ik)}\in\mathbf{B}_{i}} \exp{(\mathrm{sim}(S_{l}^{(ij)},S_{l^{\prime}}^{(ik)})/T)}} \\ + \log\frac{\exp{(\mathrm{sim}(S_{l}^{(ij)},S_{l^{\prime}}^{(ij)})/T)}}{\displaystyle\sum_{S^{(ik)}\in\mathbf{B}_{i}} \exp{(\mathrm{sim}(S_{l}^{(ik)},S_{l^{\prime}}^{(ij)})/T)}}\Bigr)
\end{split}
\label{eq:cntrs}
\end{equation}
where $T$ denotes a temperature hyperparameter to scale the cosine similarity. $\mathrm{sim}(S_{l},S_{l^{\prime}})$ is defined as:
\begin{equation}
    \mathrm{sim}(S_{l},S_{l^{\prime}}) = \mathrm{cos}(\mathbf{h}(S_{l}), \mathbf{h}(S_{l^{\prime}}))
\end{equation}
\begin{equation}
    \mathbf{h}(S_{l}) = \mathbf{W}_{1}\sigma_{cntrs}(\mathbf{W}_{2}\mathbf{u} + \mathbf{b}_{2}) + \mathbf{b}_{1}
\end{equation}
\begin{equation}
    \mathbf{h}(S_{l^{\prime}}) = \mathbf{W}_{1}\sigma_{cntrs}(\mathbf{W}_{2}\mathbf{v} + \mathbf{b}_{2}) + \mathbf{b}_{1}
\end{equation}
where $\mathbf{W}_{1} \in \mathbb{R}^{d_{cntrs}\times d}$ and $\mathbf{W}_{2} \in \mathbb{R}^{d\times d}$ mean the weights of two fully-connected layers, $\mathbf{b}_{1} \in \mathbb{R}^{d_{cntrs}}$ and $\mathbf{b}_{2} \in \mathbb{R}^{d}$ mean the biases of two fully-connected layers, and $d_{cntrs}<d$. According to~\cite{DBLP:conf/icml/ChenK0H20}, we use $\mathrm{ReLU}$~\cite{DBLP:conf/icml/NairH10} for $\sigma_{cntrs}$. Our proposed objective diverges from LaBSE's AMS loss by omitting the additive margin, incorporating MLP layers prior to loss computation, and constructing the loss in a contrastive manner. We showcase the enhanced effectiveness of this approach for MSE learning in Section~\ref{sec:5}.

In~\cite{DBLP:conf/acl/MaoGCJK20}, moreover, we introduced a sentence similarity-based contrastive task. We discard that objective in this study because we found that it has minimal impact on multilingual model training of EMS. This may be because it relies on high-dimensional sentence embeddings (e.g., 1,024) to determine similarities, while the sentence embedding dimension is reduced to a low-dimensional size in the current model architecture after adding two fully-connected layers.

\subsection{Joint Training}
We train the model by jointly optimizing the losses of the proposed generative and contrastive objectives. Specifically, we simultaneously train each batch with Eqs. (\ref{eq:xtr}) and (\ref{eq:cntrs}):
\begin{equation}
    \mathcal{L}^{(i)}=\frac{1}{\parallel\mathbf{B}_{i}\parallel}(\mathcal{L}_{XTR}^{(i)}+\mathcal{L}_{cntrs}^{(i)})
\end{equation}
where $\parallel\mathbf{B}_{i}\parallel$ denotes the number of sentence pairs within batch $\mathbf{B}_{i}$, namely, the batch size. Both $\mathcal{L}_{XTR}$ and $\mathcal{L}_{cntrs}$ play a dominant role in massively MSE training (details are given in Section~\ref{sec:5.7}).


\section{Model Training}
\label{sec:4}
In this section, we introduce the parallel corpora that we used to train language-agnostic sentence representations and specific preprocessing and training details.

\begin{table*}[t]
    \centering
    \caption{\textbf{Number of parallel sentences in each language-used model for training.} \textbf{Bold} denotes fewer data used for training. Compared with LASER, we use ~60\% of the training data in total, and we use significantly fewer parallel sentences for 43 out of 61 language pairs. The total amount of the LASER training data is calculated in these 61 languages.}
    \label{data}
    \resizebox{\linewidth}{!}{
    \begin{tabular}{l|rrrrrrrrrrrrrrrr}
        \hline
        Model & af & ar & bg & bn & ca & cs & da & de & el & eo & es & et & eu & fa & fi & fr \\
        \hline
        LASER & 67k & 8.2M & 4.9M & 913k & \textbf{813k} & 5.5M & 7.9M & 8.7M & 6.5M & \textbf{397k} & \textbf{4.8M} & 5.3M & 1.2M & - & 7.9M & 8.8M \\
        EMS (ours) & \textbf{50k} & \textbf{4.9M} & \textbf{2.8M} & \textbf{606k} & 1.0M & \textbf{3.3M} & \textbf{4.3M} & \textbf{5.6M} & \textbf{3.9M} & 683k & 9.5M & \textbf{2.7M} & \textbf{818k} & 5.1M & \textbf{4.2M} & \textbf{8.7M} \\
        \hline
        Model & gl & gu & he & hi & hr & hu & hy & id & it & ja & jv & ka & kk & ko & ku & lt \\
        \hline
        LASER & \textbf{349k} & - & 4.1M & 288k & 4M & 5.3M & \textbf{6k} & 4.3M & 8.3M & 3.2M & - & 296k & \textbf{4k} & \textbf{1.4M} & 50k & 3.2M \\
        EMS (ours) & 409k & 0.3k & \textbf{2.7M} & \textbf{199k} & \textbf{2.3M} & \textbf{3.2M} & 42k & \textbf{2.6M} & \textbf{6.1M} & \textbf{2.9M} & 0.9k & \textbf{229k} & 24k & 1.9M & \textbf{0.3k} & \textbf{2.2M} \\
        \hline
        Model & lv & mk & ml & mn & mr & ms & my & nb & nl & pl & pt & ro & ru & sk & sl & sq \\
        \hline
        LASER & 2M & 4.2M & \textbf{373k} & - & \textbf{31k} & 2.9M & \textbf{2k} & 4.1M & 8.4M & 5.5M & 8.3M & 4.9M & 9.3M & 5.2M & 5.2M & 3.2M \\
        EMS (ours) & \textbf{1.2M} & \textbf{2.4M} & 402k & 26k & 126k & \textbf{1.9M} & 3k & \textbf{46k} & \textbf{4.8M} & \textbf{3.2M} & \textbf{6.1M} & \textbf{3.0M} & \textbf{6.2M} & \textbf{2.8M} & \textbf{2.8M} & \textbf{2.1M} \\
        \hline
        Model & sr & sv & sw & ta & te & th & tl & tr & uk & ur & vi & yo & zh & & \multicolumn{2}{|c}{Total} \\
        \hline
        LASER & 4M & 7.8M & 173k & 42k & 33k & 4.1M & \textbf{36k} & 5.7M & \textbf{1.4M} & 746k & 4M & - & 8.3M & & \multicolumn{2}{|c}{204M} \\
        EMS (ours) & \textbf{2.4M} & \textbf{4.2M} & \textbf{41k} & 42k & \textbf{30k} & \textbf{2.2M} & 45k & \textbf{3.8M} & 1.5M & \textbf{50k} & \textbf{2.6M} & 0.2k & \textbf{6.6M} & & \multicolumn{2}{|c}{\textbf{143M}} \\
        \hline
    \end{tabular}
    }
\end{table*}

\subsection{Training Data}
\label{sec:4.1}
We collected parallel corpora for 62 languages from OPUS\footnote{\url{https://opus.nlpl.eu/}}~\cite{DBLP:conf/lrec/Tiedemann12} (See Table~\ref{data}).\footnote{We do not distinguish between traditional and simplified Chinese.} The 62 languages that we selected cover all the languages in~\cite{DBLP:conf/eacl/SchwenkCSGG21} and the languages suggested by the cross-lingual generalization benchmark, XTREME~\cite{DBLP:conf/icml/HuRSNFJ20}. While gathering each corpus, we used toolkits provided by~\cite{DBLP:conf/lrec/AulamoSVT20}\footnote{\url{https://github.com/Helsinki-NLP/OpusTools}} and~\cite{DBLP:conf/emnlp/ReimersG20}.\footnote{\url{https://github.com/UKPLab/sentence-transformers}} Specifically, we used the following corpora for training:

\noindent
\textbf{Europarl} is a parallel corpus extracted from the European Parliament website by Philipp Koehn~\cite{DBLP:conf/mtsummit/Koehn05}. We used the entire corpus for each language pair.

\noindent
\textbf{GlobalVoices} is a parallel corpus of news stories from the website Global Voices compiled and provided by CASMACAT.\footnote{\url{http://casmacat.eu/corpus/global-voices.html}} We used the entire corpus for each language pair.

\noindent
\textbf{NewsCommentary} is a news commentary parallel corpus provided by WMT\footnote{\url{https://statmt.org/}} for training statistical machine translation. We used the entire corpus for each language pair.

\noindent
\textbf{OpenSubtitles} is a parallel corpus of movie subtitles collected from \url{opensubtitles.org}~\cite{DBLP:conf/lrec/LisonT16}. Considering that the lengths of most sentences are short, we used at most 2M sentence pairs for each language pair to control the training data size.

\noindent
\textbf{Ted} is a parallel corpus comprising TED talks. We used the 2020 version crawled by~\cite{DBLP:conf/emnlp/ReimersG20}, which includes ~4000 TED talks for each language pair available.

\noindent
\textbf{UNPC} United nations parallel corpus of six languages~\cite{DBLP:conf/lrec/ZiemskiJP16}. We used 5M sentence pairs for en--ru and 2M sentence pairs for other language pairs.\footnote{As the number of en--ru sentence pairs from other parallel corpora is relatively small, we used more for data for en--ru to balance the size for different language pairs.}

\noindent
\textbf{WikiMatrix} is a parallel corpus crawled by~\cite{DBLP:conf/eacl/SchwenkCSGG21}. We used the entire corpus for each language pair.

\noindent
\textbf{Tatoeba} is a parallel corpus gathered from Tatoeba's website,\footnote{https://tatoeba.org/} the language learning supporting website. As training on the Tatoeba benchmark will probably improve the evaluation performance on the Tatoeba benchmark~\cite{DBLP:journals/tacl/ArtetxeS19}, following~\cite{DBLP:conf/emnlp/ReimersG20}, we excluded the training data of Tatoeba for most language pairs. Only for the language pairs that are not included in the aforementioned corpora, we used Tatoeba corpora.

The aforementioned training data leads to a 143M parallel corpus. As listed in Table~\ref{data}, we used much fewer data for 43 languages than LASER. Moreover, we excluded the JW300~\cite{DBLP:conf/acl/AgicV19} corpus and pruned OpenSubtitles and UNPC corpora and included less training data than SBERT-distill.\footnote{\cite{DBLP:conf/emnlp/ReimersG20} used JW300 and all of the entire corpora we used.} In the next section (Section\ref{sec:5}), we will show that our model yields better or comparable sentence representation performance, compared with LASER and SBERT-distill. Considering our model's ability to deliver superior outcomes with reduced training data, it becomes feasible to extend our model to accommodate more low-resource languages, even with limited data availability. 

\subsection{Preprocessing Details}
For the parallel corpus containing 62 languages, we removed the sentences that appear in any evaluation dataset (see Section~\ref{sec:5}). We tokenized Chinese using jieba\footnote{\url{https://github.com/fxsjy/jieba}} and Japanese using Jumanpp\footnote{\url{https://github.com/ku-nlp/jumanpp}}~\cite{DBLP:conf/emnlp/MoritaKK15,DBLP:conf/emnlp/TolmachevKK18}, as the application of language-specific word segmentation for Japanese and Chinese has been shown to enhance performance across various tasks, including NMT~\cite{DBLP:conf/acl/ProvilkovEV20,DBLP:conf/acl/Kudo18,DBLP:conf/lrec/MaoCDSK20,DBLP:journals/talip/MaoCK22} and MLM pre-training~\cite{DBLP:conf/nips/ConneauL19,DBLP:conf/ijcnlp/XuGLW20}. We used Moses tokenizer for other languages.\footnote{\url{https://github.com/moses-smt/mosesdecoder/blob/master/scripts/tokenizer/tokenizer.perl}} We converted all the sentences to lowercase. Subsequently, we applied SentencePiece\footnote{\url{https://github.com/google/sentencepiece}}~\cite{DBLP:conf/emnlp/KudoR18} to convert words to subwords, which leads to a vocabulary with 60k tokens.\footnote{SBERT-distill used a vocabulary of 250k tokens, which significantly improved the model parameters. In contrast, we used 60k, which is comparable with LASER's 50k vocabulary.} Finally, we add 62 language tokens (e.g., $<2en>$, $<2fr>$, ...) to the 60k vocabulary.

\begin{table}[t]
    \centering
    \caption{\textbf{Values of the hyperparameters tuned by grid search.} \textbf{Bold} denotes the best hyperparameter combination.}
    \label{hyper}
    \begin{tabular}{ l r }
        \hline
        Hyperparameters & Values \\
        \hline
        number of the transformer layers & 2, 4, \textbf{6}, 12 \\
        transformer hidden dropout & 0.0, \textbf{0.1}, 0.3 \\
        transformer attention dropout & 0.0, \textbf{0.1} \\
        $T$ & 0.01, \textbf{0.1}, 0.2, 0.5, 1.0 \\
        learning rate & 1e-4, \textbf{3e-4}, 5e-4, 1e-3 \\
        weight decay & 0.0, \textbf{1e-5}, 1e-4, 1e-3 \\
        warm-up steps & 0, 5,000, \textbf{10,000}, 20,000 \\
        \hline
    \end{tabular}
\end{table}

\subsection{Training Details}
We employed transformer encoder~\cite{DBLP:conf/nips/VaswaniSPUJGKP17} as the basic unit of the training architecture (Fig.~\ref{architecture}). We conducted a grid search for optimal hyperparameter combinations by observing the validation loss on the WikiMatrix validation datasets (Table~\ref{hyper}).

As a result, the dual transformer encoder sharing parameters has 6 layers, 16 attention heads, a hidden size of 1,024, and a feed-forward size of 4,096. The transformer encoder can be substituted by encoders with other structures. $d$, $d_{vcb}$, $d_{la}$, and $d_{cntrs}$ are 1,024, 60,000, 128, and 128, respectively. We set 0.1 for the temperature $T$ of the contrastive objective.

For the model training, we fed the parallel sentences into the dual transformer encoder and truncated the sentences up to 120 tokens.\footnote{Although LASER and SBERT-distill allowed much longer sentences during the training phase, we demonstrate that 120 tokens are sufficient for a single sentence with complete semantics. For the evaluation, documents longer than 120 tokens can be separated into several sentences, which would not limit the usage of our model.} We trained three epochs for the entire training corpus with the Adam optimizer~\cite{DBLP:journals/corr/KingmaB14}, the learning rate of 0.0003 with the linear warm-up strategy of 10,000 steps, a weight decay of 0.00001, and a dropout\footnote{the hidden and attention dropouts} of 0.1 for the transformer encoder. We used four V100 GPUs to conduct the model training with a batch size of 152 parallel sentences.

\begin{table}[t]
    \centering
    \caption{\textbf{Comparison between related studies and the proposed EMS.} ``\#Langs'' and ``\#Paral'' denote the number of languages the model supports and the number of parallel sentences used for training, respectively. ``Mono'' means whether the model incorporated monolingual data for training; ``Archit.'' denotes the model architecture; ``\#Param.'' indicates the number of model parameters.} 
    \label{models}
    \resizebox{\linewidth}{!}{
    \begin{tabular}{ l r r r r r}
        \hline
        Model & \#Langs & \#Paral & Mono & Archit. & \#Param. \\
        \hline
        LASER & 93 & 223M & & Seq2seq-LSTM & 148M \\
        SBERT-distill & 50 & $>>$223M & $\surd$ & Dual-Trans &  270M \\
        LaBSE & 109 & 6B & $\surd$ & Dual-Trans & 471M \\
        EMS (ours) & 62 & 143M & & Dual-Trans & 147M \\
        \hline
    \end{tabular}
    }
\end{table}

\subsection{Efficiency Comparison with Competing Models}

The superior efficiency of EMS stems from two key aspects: data efficiency and computation efficiency. In terms of data efficiency, as listed in Table~\ref{models}, the proposed method includes 143M parallel sentences for model training, which is significantly less than those of other massive MSE models. Nevertheless, as demonstrated in Section~\ref{sec:5}, the reduced data for EMS results in a comparable or even improved model.


In terms of training efficiency, we employed the encoder-only architecture as the basic model unit, whereas LASER required the encoder-decoder architecture to perform the translation task, where the presence of the decoder decreased training efficiency. LASER2 utilized Transformer architecture to streamline the LSTM-based LASER approach. However, as indicated in Section~\ref{sec:5}, LASER2 falls short in MSE construction compared to LASER. Conversely, our proposed EMS, leveraging the XTR objective, could capture finer-grained information comparable to LASER's full sentence generation, thus enhancing performance and streamlining the LASER framework. The above indicates that the proposed model is an alternative to LASER, whereas SBERT-distill and LaBSE are complementary because the distillation from the English-SBERT and the use of the pre-trained model of LaBSE are feasible to be combined with the proposed training objectives. By leveraging a pre-trained model for initialization, EMS could potentially see further improvements. In this study, we do not use pre-trained models for comparison, as LaBSE's pre-trained model is not publicly accessible. Instead, we integrate LaBSE's AMS loss into our EMS framework, maintaining consistency in training data and model architecture, to determine the most efficient and effective training objective. For detailed configurations and results, please refer to Section~\ref{sec:5}.

With regard to the specific training time, the loss nearly converged after being trained for 0.5 epochs (122,196 steps) and converged completely after 3 epochs (733,176 steps), whereas LASER is trained for 17 epochs till convergence. Concerning the training time, SBERT-distill and LaBSE rely on large-scale pre-trained models; thus, their fine-tuning requires heavy computation as operating forwarding on large-scale models. LASER is trained with 80 V100 GPU$\times$days, while our EMS requires 5  V100 GPU$\times$days to nearly converge and 20 V100 GPU$\times$days to converge fully, which indicates 4$\sim$16 times speedup of EMS compared with LASER. In Section~\ref{sec:5.7}, we delve deeper into the training efficiency of each model component, demonstrating that the proposed generative and contrastive objectives, along with the language embedding and MLP layers for the contrastive objective, can be implemented efficiently without significantly increasing computational demands in a dual-encoder architecture.

In terms of inference efficiency, which can vary significantly depending on computational resources, we do not report the absolute time required for inference. As indicated in Table~\ref{models}, our model has the fewest parameters, resulting in the quickest embedding inference time compared to LASER, SBERT-distill, and LaBSE. However, this does not lead to faster application in downstream tasks, as the embedding dimension remains at 1,024, identical to LASER and larger than 768 of both SBERT-distill and LaBSE. To enhance efficiency in downstream applications, the distillation technique from LEALLA~\cite{DBLP:conf/eacl/MaoN23} could be employed to create a lower-dimensional version of EMS with comparable performance.


In addition, other previous studies on learning language-agnostic sentence representation models, e.g.,~\cite{DBLP:conf/wmt/GuoSYGCASCSSK18,DBLP:conf/rep4nlp/ChidambaramYCYS19,DBLP:conf/ijcai/YangAYGSCSSK19}, and mUSE~\cite{DBLP:conf/acl/YangCAGLCAYTSSK20}, propose the training objective of distinguishing the positive translation from several \textit{hard negative} samples. The heavy computation load of \textit{hard negative} samples for each sentence limits the feasibility of their methods to a small number of languages, i.e., fewer than 16.


\section{Evaluation}
\label{sec:5}

In this section, we evaluate the performance of the language-agnostic sentence representation on two groups of downstream tasks. On the one hand, without any further fine-tuning, we test the parallel sentence retrieval capability of the model using the cosine similarity between sentences. We evaluate this based on the following four tasks: Tatoeba benchmark~\cite{DBLP:journals/tacl/ArtetxeS19}, Flores benchmark~\cite{DBLP:journals/tacl/GoyalGCCWJKRGF22}, BUCC benchmark~\cite{DBLP:conf/acl-bucc/ZweigenbaumSR17,zweigenbaum2018overview}, and cross-lingual sentence retrieval on the ParaCrawl corpus~\cite{DBLP:conf/acl/BanonCHHHEFKKKO20}. On the other hand, by fine-tuning a simple multi-layer perceptron, we evaluate the model performance based on three cross-lingual sentence classification tasks in a zero-shot manner. Three evaluation tasks include the MLDoc benchmark~\cite{DBLP:conf/lrec/SchwenkL18} and cross-lingual sentiment classification on two versions of the multilingual Amazon review corpora~\cite{DBLP:conf/acl/PrettenhoferS10,DBLP:conf/emnlp/KeungLSS20}. The former group of the evaluation measures the alignment performance of the language-agnostic sentence representation space, whereas the latter group evaluates the fundamental natural language classification ability of the model. More complicated cross-lingual natural language understanding (XNLU) tasks, e.g., XNLI~\cite{DBLP:conf/emnlp/ConneauRLWBSS18} and XQuAD~\cite{DBLP:conf/acl/ConneauKGCWGGOZ20}, have been comprehensively proven to perform better with cross-lingual language model pre-training and fine-tuning in XTREME~\cite{DBLP:conf/icml/HuRSNFJ20}, while \textit{fixed representation} models are not competent to address such tasks~\cite{DBLP:journals/tacl/ArtetxeS19}. Thus, we do not include the evaluation of XNLU in this study. Furthermore, we analyze the effectiveness of each component of the model structure based on an ablation study.

For all the evaluation tasks, we compare the following massively multilingual sentence representation models:

\noindent
\textbf{LASER}
\cite{DBLP:journals/tacl/ArtetxeS19} employed the BiLSTM encoder-decoder to train MSE for 93 languages by optimizing the translation task. 223M parallel sentences are used for training. 


\noindent
\textbf{SBERT-distill}
\cite{DBLP:conf/emnlp/ReimersG20} trained MSE for 50 languages by distilling the monolingual pre-trained encoder. Our training data are a subset of their data (Section~\ref{sec:4.1}). ``paraphrase-xlm-r-multilingual-v1'' is used for evaluation.\footnote{\url{https://github.com/UKPLab/sentence-transformers/tree/master/examples/training/paraphrases}}

\noindent
\textbf{\textit{LaBSE}}
\cite{DBLP:conf/acl/FengYCA022} trained MSE for 109 languages by fine-tuning the sentence-level contrastive task from mBERT. We \textit{italicize} this model in the following tables (results) as the upper bound performance on downstream tasks because a large number of parallel sentences, 6B, are used for training. LaBSE continues to be the leading state-of-the-art model for parallel sentence retrieval in a massively multilingual scenario.

\noindent
\textbf{EMS (ours)}
We trained an MSE model for 62 languages. We used significantly less training data, thus less computation overhead, than those used in the previous study. The proposed model can be easily trained from scratch with competitive MSE performance.

\noindent
\textbf{LaBSE-EMS-vanilla}
In our EMS model architecture, we implement the AMS loss of LaBSE for learning MSE, a variant of the contrastive loss originally introduced in~\cite{DBLP:conf/ijcai/YangAYGSCSSK19}. We substitute the standard contrastive loss in EMS with AMS, setting the margin to 0.3 and scaling temperature to 0.1 without MLP layers upon sentence embedding and the XTR objective.

\noindent
\textbf{LaBSE-EMS-MLP}
In this setting, based on LaBSE-EMS-vanilla, AMS is further applied with MLP layers on top of the sentence embeddings, as introduced in Section~\ref{sec:3.3}.

\noindent
\textbf{LaBSE-EMS-joint (ours)}
This setting evaluates the compatibility of the MLP-based AMS contrastive loss with our XTR generative loss within the EMS framework, combining LaBSE-EMS-MLP and XTR.

LaBSE-EMS-vanilla, LaBSE-EMS-MLP, and LaBSE-EMS-joint (ours) are for identifying the most effective contrastive loss for MSE and to assess the compatibility of LaBSE's AMS with EMS's XTR objectives. We also include the results of \textbf{LASER2}~\cite{DBLP:conf/emnlp/HeffernanCS22} on Tatoeba\footnote{As LASER2 only outperforms LASER in a handful of low-resource languages (See Section~\ref{sec:5.1}), we solely present the findings on Tatoeba.}, \textbf{mUSE}~\cite{DBLP:conf/acl/YangCAGLCAYTSSK20} on Tatoeba and BUCC, and \textbf{LaBSE-bilingual}~\cite{DBLP:conf/ijcai/YangAYGSCSSK19} on BUCC benchmarks for comparison.\footnote{mUSE serves as a crucial baseline on Tatoeba and BUCC for high-resource languages that have been utilized in previous studies~\cite{DBLP:conf/emnlp/ReimersG20,DBLP:conf/acl/FengYCA022,DBLP:conf/acl/MaoGCJK20}.} LASER2 is a more recent version of the LASER model, incorporating SentencePiece~\cite{DBLP:conf/emnlp/KudoR18} instead of BPE~\cite{DBLP:conf/acl/SennrichHB16a}, trained with transformer, and can yield improved results for low-resource languages. On the other hand, mUSE is a universal sentence encoder model that is only compatible with 16 languages. In this study, we do not include LASER3-related models~\cite{DBLP:conf/emnlp/HeffernanCS22,DBLP:conf/eacl/TanHSK23} in our analysis, as their focus on adapting existing models to low-resource languages by language-specific parameters diverges from our objective of identifying efficient and effective objectives for one-for-all models.

\begin{table*}[t!]
    \centering
    \caption{\textbf{P@1 results on Tatoeba benchmark.} \textbf{Bold} fonts denote the best precisions among all the models except LaBSE. We report the average precision of the English$\rightarrow$X and X$\rightarrow$English for each language.}
    \label{tatoeba}
    \resizebox{0.90\linewidth}{!}{
    \begin{tabular}{l|rrrrrrrrrrrrrrrr}
        \hline
        Model & afr & ara & ben & bul & cat & ces & cmn & dan & deu & ell & epo & est & eus & fin & fra \\
        \hline
        LASER & 89.5 & 92.0 & \textbf{89.6} & 95.0 & 95.9 & 96.5 & 95.4 & 96.0 & 99.0 & 95.0 & 97.2 & 96.7 & 94.6 & 96.3 & 95.6 \\
        LASER2 & 85.5 & 85.8 & 87.2 & 90.3 & 92.4 & 93.1 & 69.2 & 90.3 & 93.7 & 93.2 & 90.5 & 93.2 & 85.2 & 89.3 & 92.0 \\
        SBERT-distill & 84.5 & 87.7 & 77.6 & 94.0 & 96.4 & 96.3 & 95.0 & 96.2 & 98.7 & 95.5 & 68.8 & 95.8 & 48.6 & 96.4 & 94.7 \\
        LaBSE-EMS-vanilla & 79.6 & 84.3 & 82.9 & 86.3 & 92.7 & 93.9 & 90.2 & 93.2 & 97.5 & 91.0 & 96.2 & 93.3 & 89.3 & 93.3 & 93.0 \\
        LaBSE-EMS-MLP & 90.3 & 89.7 & 88.0 & 91.4 & 96.0 & 96.3 & 94.3 & 95.1 & 99.4 & 95.4 & 98.0 & 96.6 & 93.6 & 96.1 & 95.6 \\
        LaBSE-EMS-joint (ours) & 93.1 & 93.0 & 89.5 & 95.6 & \textbf{97.2} & \textbf{97.5} & 95.4 & 96.3 & 99.1 & \textbf{96.5} & 98.6 & 97.7 & \textbf{95.1} & 97.1 & 96.3 \\
        EMS (ours) & \textbf{94.0} & \textbf{93.9} & 83.8 & \textbf{95.8} & 97.0 & 97.4 & \textbf{95.9} & \textbf{97.0} & \textbf{99.3} & \textbf{96.5} & \textbf{98.9} & \textbf{97.8} & 94.9 & \textbf{98.0} & 96.2 \\
        \hline
        \textit{LaBSE} & \textit{97.4} & \textit{91.0} & \textit{91.3} & \textit{95.7} & \textit{96.5} & \textit{97.5} & \textit{96.2} & \textit{96.4} & \textit{99.4} & \textit{96.6} & \textit{98.4} & \textit{97.7} & \textit{95.8} & \textit{97.0} & \textit{96.0} \\
        \hline
        \hline
        Model & glg & heb & hin & hrv & hun & hye & ind & ita & jav & jpn & kat & kaz & kor & kur & lit \\
        \hline
        LASER & 95.5 & 92.2 & 94.7 & 97.2 & 96.0 & 36.1 & 94.5 & 95.3 & 22.9 & 90.7 & 35.9 & 18.6 & 88.9 & 17.2 & 96.2 \\
        LASER2 & 88.9 & 84.6 & 88.3 & 94.1 & 90.5 & 81.8 & 88.1 & 92.3 & 17.1 & 88.2 & 69.6 & 38.4 & 77.7 & 12.1 & 92.9 \\
        SBERT-distill & 96.0 & 88.4 & \textbf{96.4} & 97.0 & 94.7 & \textbf{91.3} & 94.1 & 94.9 & 37.3 & 94.2 & \textbf{91.4} & \textbf{73.7} & 90.1 & \textbf{43.7} & 95.8 \\
        LaBSE-EMS-vanilla & 90.6 & 84.0 & 86.8 & 94.0 & 92.7 & 74.8 & 91.4 & 90.5 & 45.1 & 89.1 & 56.1 & 50.5 & 84.4 & 18.7 & 93.5 \\
        LaBSE-EMS-MLP & 95.6 & 89.6 & 92.5 & 96.8 & 96.6 & 81.2 & 94.4 & 93.5 & 43.9 & 93.7 & 64.1 & 55.7 & 88.4 & 22.3 & 96.1 \\
        LaBSE-EMS-joint (ours) & 96.9 & 92.2 & 95.2 & \textbf{97.5} & 97.6 & 85.0 & 95.8 & 96.3 & 49.0 & 95.3 & 70.1 & 60.3 & 90.8 & 27.4 & 96.8 \\
        EMS (ours) & \textbf{97.1} & \textbf{92.5} & 93.4 & \textbf{97.5} & 97.4 & 87.8 & \textbf{95.8} & \textbf{96.7} & \textbf{55.6} & \textbf{95.8} & 73.5 & 63.8 & \textbf{92.3} & 31.3 & \textbf{97.3} \\
        \hline
        \textit{LaBSE} & \textit{97.2} & \textit{93.0} & \textit{97.7} & \textit{97.8} & \textit{97.2} & \textit{95.0} & \textit{95.3} & \textit{94.6} & \textit{84.4} & \textit{96.4} & \textit{95.9} & \textit{90.5} & \textit{93.5} & \textit{87.1} & \textit{97.3} \\
        \hline
        \hline
        Model & lvs & mal & mar & max & mkd & mon & nld & nob & pes & pol & por & ron & rus & slk & slv \\
        \hline
        LASER & 95.4 & 96.9 & 91.5 & 50.9 & 94.7 & 8.2 & 96.3 & \textbf{98.8} & 93.4 & 97.8 & 95.2 & 97.4 & 94.6 & 96.6 & 95.9 \\
        LASER2 & 92.2 & 95.1 & 89.5 & 30.3 & 89.2 & 2.8 & 92.4 & 86.9 & 84.1 & 91.2 & 92.4 & 93.0 & 91.2 & 93.9 & 91.1 \\
        SBERT-distill & 96.4 & 94.0 & 91.0 & 58.5 & 92.2 & \textbf{91.7} & 96.0 & 98.0 & 94.8 & 97.0 & 94.8 & 96.4 & 93.5 & 96.2 & 95.5 \\
        LaBSE-EMS-vanilla & 91.8 & 90.3 & 88.8 & 58.1 & 87.6 & 59.0 & 92.4 & 91.6 & 90.8 & 94.7 & 92.4 & 94.4 & 92.5 & 93.9 & 93.4 \\
        LaBSE-EMS-MLP & 94.9 & 95.7 & 90.8 & 60.4 & 93.6 & 68.4 & 96.5 & 95.5 & 94.7 & 96.7 & 94.5 & 96.8 & 94.7 & 96.7 & 95.3 \\
        LaBSE-EMS-joint (ours) & 96.7 & \textbf{97.1} & \textbf{94.3} & 66.2 & 96.4 & 72.0 & \textbf{97.7} & 97.4 & 95.6 & 98.1 & \textbf{96.1} & \textbf{98.0} & \textbf{95.6} & \textbf{97.7} & 96.7 \\
        EMS (ours) & \textbf{96.9} & 77.8 & 88.2 & \textbf{69.9} & \textbf{97.0} & 73.9 & \textbf{97.7} & 97.5 & \textbf{96.0} & \textbf{98.2} & 95.9 & 97.9 & 95.2 & 97.5 & \textbf{97.1} \\
        \hline
        \textit{LaBSE} & \textit{96.8} & \textit{98.9} & \textit{94.8} & \textit{71.1} & \textit{94.8} & \textit{96.6} & \textit{97.2} & \textit{98.9} & \textit{96.0} & \textit{97.8} & \textit{95.6} & \textit{97.8} & \textit{95.3} & \textit{97.3} & \textit{96.7} \\
        \hline
        \hline
        Model & spa & sqi & srp & swe & swh & tam & tel & tgl & tha & tur & ukr & urd & vie & \multicolumn{2}{|c}{Avg.} \\
        \hline
        LASER & 98.0 & 98.0 & 95.3 & 96.6 & \textbf{57.6} & 69.4 & 79.7 & 50.6 & 95.4 & 97.5 & 94.5 & 81.9 & 96.8 & \multicolumn{2}{|c}{84.7} \\
        LASER2 & 93.4 & 94.9 & 89.5 & 92.1 & 44.4 & 77.9 & \textbf{93.6} & 50.1 & 92.1 & 95.3 & 91.5 & 71.9 & 89.9 & \multicolumn{2}{|c}{81.5} \\
        SBERT-distill & 98.0 & 97.5 & 93.8 & 95.7 & 27.6 & \textbf{85.7} & 89.1 & 32.4 & 96.3 & 97.2 & 94.3 & \textbf{92.2} & 97.2 & \multicolumn{2}{|c}{87.7} \\
        LaBSE-EMS-vanilla & 93.8 & 95.7 & 92.4 & 92.3 & 32.8 & 65.5 & 57.1 & 70.1 & 93.2 & 94.9 & 92.5 & 69.0 & 92.7 & \multicolumn{2}{|c}{83.7} \\
        LaBSE-EMS-MLP & 98.5 & 97.9 & 95.8 & 95.4 & 37.1 & 69.7 & 64.1 & 76.6 & 95.0 & 97.8 & 94.7 & 76.3 & 96.5 & \multicolumn{2}{|c}{87.6} \\
        LaBSE-EMS-joint (ours) & \textbf{98.7} & \textbf{98.4} & 96.0 & 96.5 & 45.9 & 74.6 & 74.8 & 82.1 & 96.8 & \textbf{99.0} & \textbf{96.0} & 84.2 & 97.5 & \multicolumn{2}{|c}{\textbf{90.0}} \\
        EMS (ours) & 98.6 & \textbf{98.4} & \textbf{96.4} & \textbf{97.0} & 53.2 & 52.8 & 69.7 & \textbf{84.8} & \textbf{97.4} & 98.6 & \textbf{96.0} & 86.0 & \textbf{97.7} & \multicolumn{2}{|c}{89.8} \\
        \hline
        \textit{LaBSE} & \textit{98.4} & \textit{97.6} & \textit{96.2} & \textit{96.5} & \textit{88.6} & \textit{90.7} & \textit{98.3} & \textit{97.4} & \textit{97.1} & \textit{98.4} & \textit{95.2} & \textit{95.3} & \textit{97.8} & \multicolumn{2}{|c}{\textit{95.3}} \\
        \hline
    \end{tabular}
    }
\end{table*}

\begin{table*}[t]
    \centering
    \caption{\textbf{Average P@1 results of different groups of the languages on Tatoeba benchmark.} \textbf{Bold} are the best precisions among all the models except LaBSE. ``mUSE,'' ``XTREME,'' and ``SBERT-distill'' denote the 15, 38, and 48 languages that the respective model or benchmark includes. ``$<$LASER'' denotes the 43 languages that use less training data than LASER. ``$>$300k'' and ``$<$300k'' indicate that LASER, LASER2, and EMS (the proposed model) include more than or less than 300k parallel sentences for training. Refer to Table~\ref{data}; ``$>$300k'' and ``$<$300k'' contain 42 and 11 languages, respectively.}
    \label{tatoeba-all}
    \resizebox{0.80\linewidth}{!}{
    \begin{tabular}{l|rrrrrrr}
        \hline
        Model & mUSE (15) & XTREME (38) & SBERT-distill (48) & $<$LASER (43) & $>$300k (42) & $<$300k (11) \\
        \hline
        mUSE & 93.9 & - & - & - & - & - \\
        LASER & 95.1 & 84.2 & - & 89.6 & 94.4 & 58.3 \\
        LASER2 & 89.0 & 80.6 & - & 85.0 & 88.9 & 66.4 \\
        SBERT-distill & 94.9 & 85.5 & 94.8 & - & 92.1 & \textbf{73.3} \\
        LaBSE-EMS-vanilla & 91.6 & 81.8 & 89.3 & 85.3 & 90.9 & 61.9 \\
        LaBSE-EMS-MLP & 94.9 & 86.0 & 93.0 & 89.1 & 94.2 & 67.7 \\
        LaBSE-EMS-joint (ours) & 96.3 & \textbf{88.8} & 94.8 & 91.4 & \textbf{95.7} & 73.0 \\
        EMS (ours) & \textbf{96.6} & 88.2 & \textbf{95.0} & \textbf{91.8} & 95.4 & 72.0 \\
        \hline
        \textit{LaBSE} & \textit{96.2} & \textit{94.7} & - & - & \textit{95.8} & \textit{93.9} \\
        \hline
    \end{tabular}
    }
\end{table*}

\begin{table*}[t]
    \centering
    \caption{\textbf{P@1 results of EMS's unseen languages on the Tatoeba benchmark.} Bold fonts denote the best precisions among all the models except LaBSE. Several languages are trained in LASER and LaBSE. We report the average precision of the English$\rightarrow$X and X$\rightarrow$English for each language.}
    \label{unseen}
    \resizebox{0.85\linewidth}{!}{
    \begin{tabular}{l|rrrrrrrrrrrrrrr}
        \hline
        Model & amh & ang & arq & arz & ast & awa & aze & bel & ber & bos & bre & cbk & ceb & cha \\
        \hline
        LASER & 42.0 & 37.7 & 39.5 & 68.9 & 86.2 & 36.1 & 66.0 & 69.6 & 68.2 & 96.5 & 15.8 & 77.0 & 15.7 & 29.2 \\
        LASER2 & \textbf{69.4} & 14.2 & 22.5 & 53.7 & 68.9 & 24.7 & 63.4 & 56.4 & \textbf{72.0} & 95.1 & \textbf{23.4} & 57.8 & 6.6 & 12.0 \\
        SBERT-distill & 67.9 & 25.0 & 30.6 & 63.7 & 78.3 & 46.5 & \textbf{85.0} & \textbf{86.9} & 6.8 & 95.8 & 10.1 & 69.4 & 11.7 & 25.9 \\
        LaBSE-EMS-vanilla & 3.3 & 29.9 & 30.5 & 61.4 & 80.7 & 45.0 & 40.0 & 45.2 & 5.2 & 93.9 & 6.8 & 60.5 & 19.1 & 28.8 \\
        LaBSE-EMS-MLP & 2.1 & 32.5 & 33.9 & 66.4 & 82.7 & 48.3 & 49.1 & 51.5 & 4.8 & 95.9 & 6.5 & 69.8 & 22.6 & 29.9 \\
        LaBSE-EMS-joint (ours) & 2.4 & 43.7 & 46.0 & 76.6 & \textbf{89.0} & 56.5 & 58.8 & 66.0 & 7.3 & \textbf{96.6} & 11.3 & \textbf{80.6} & 27.1 & 40.9 \\
        EMS (ours) & 0.6 & \textbf{47.4} & \textbf{48.7} & \textbf{77.7} & 88.2 & 56.1 & 62.1 & 70.3 & 7.6 & \textbf{96.6} & 12.0 & \textbf{80.6} & \textbf{30.1} & \textbf{46.4} \\
        \hline
        \textit{LaBSE} & \textit{94.0} & \textit{64.2} & \textit{46.2} & \textit{78.4} & \textit{90.6} & \textit{73.2} & \textit{96.1} & \textit{96.2} & \textit{10.4} & \textit{96.2} & \textit{17.3} & \textit{82.5} & \textit{70.9} & \textit{39.8} \\
        \hline
        \hline
        Model & cor & csb & cym & dsb & dtp & fao & fry & gla & gle & gsw & hsb & ido & ile & ina \\
        \hline
        LASER & 7.5 & 43.3 & 8.6 & 48.0 & 7.2 & 71.6 & 51.7 & 3.7 & 5.2 & 44.4 & 54.5 & 83.7 & 86.2 & \textbf{95.2} \\
        LASER2 & 4.9 & 23.9 & 5.9 & 37.2 & 4.0 & 49.1 & 34.4 & 2.1 & 3.8 & 28.6 & 38.6 & 66.4 & 82.0 & 85.1 \\
        SBERT-distill & 5.1 & 40.5 & \textbf{34.9} & 51.9 & 7.3 & \textbf{50.8} & 58.4 & \textbf{7.5} & \textbf{18.6} & 36.8 & 57.6 & 56.0 & 70.5 & 87.9 \\
        LaBSE-EMS-vanilla & 4.4 & 46.6 & 9.4 & 47.0 & 7.1 & 27.3 & 50.9 & 3.6 & 3.8 & 36.8 & 52.9 & 67.5 & 58.7 & 80.9 \\
        LaBSE-EMS-MLP & 4.5 & 50.8 & 9.5 & 50.3 & 6.2 & 30.3 & 55.8 & 3.4 & 4.2 & 39.3 & 58.5 & 77.5 & 76.6 & 88.2 \\
        LaBSE-EMS-joint (ours) & 6.5 & 64.2 & 13.7 & 65.0 & 8.2 & 39.1 & \textbf{65.6} & 6.0 & 7.8 & 50.0 & 73.0 & 84.5 & 82.3 & 93.4 \\
        EMS (ours) & \textbf{7.8} & \textbf{69.2} & 16.3 & \textbf{69.7} & \textbf{9.5} & 47.3 & 63.9 & 6.8 & 7.8 & \textbf{54.7} & \textbf{79.0} & \textbf{88.1} & \textbf{86.4} & 94.0 \\
        \hline
        \textit{LaBSE} & \textit{12.8} & \textit{56.1} & \textit{93.6} & \textit{69.3} & \textit{13.3} & \textit{90.6} & \textit{89.9} & \textit{88.8} & \textit{95.0} & \textit{52.1} & \textit{71.2} & \textit{90.9} & \textit{87.1} & \textit{95.8} \\
        \hline
        \hline
        Model & isl & kab & khm & kzj & lat & lfn & mhr & nds & nno & nov & oci & orv & pam & pms \\
        \hline
        LASER & \textbf{95.6} & 58.1 & 20.6 & 7.2 & \textbf{58.5} & 64.5 & 10.4 & \textbf{82.9} & 88.3 & 66.0 & 61.2 & 28.1 & 6.0 & 49.6 \\
        LASER2 & 90.8 & \textbf{60.8} & \textbf{65.4} & 2.9 & 46.5 & 44.5 & 5.4 & 64.7 & 55.5 & 53.7 & 45.4 & 19.2 & 2.8 & 28.7 \\
        SBERT-distill & 75.8 & 2.7 & 64.8 & 8.0 & 28.0 & 57.7 & 11.9 & 50.7 & \textbf{89.3} & 58.8 & 52.4 & 33.4 & 7.0 & 44.3 \\
        LaBSE-EMS-vanilla & 11.4 & 3.9 & 1.2 & 6.4 & 28.7 & 51.5 & 8.7 & 48.8 & 67.4 & 62.1 & 47.7 & 32.1 & 7.0 & 42.4 \\
        LaBSE-EMS-MLP & 11.7 & 3.6 & 1.6 & 6.2 & 28.8 & 57.6 & 9.2 & 60.4 & 75.0 & 66.9 & 54.2 & 36.7 & 6.9 & 48.4 \\
        LaBSE-EMS-joint (ours) & 17.9 & 4.2 & 1.1 & 9.2 & 43.6 & 68.1 & 14.2 & 70.7 & 81.5 & 74.3 & 64.6 & 47.1 & 12.7 & 60.0 \\
        EMS (ours) & 22.2 & 4.7 & 1.2 & \textbf{10.4} & 46.4 & \textbf{72.1} & \textbf{14.5} & 73.7 & 84.9 & \textbf{75.7} & \textbf{67.4} & \textbf{50.1} & \textbf{14.2} & \textbf{67.7} \\
        \hline
        \textit{LaBSE} & \textit{96.2} & \textit{6.2} & \textit{83.2} & \textit{14.2} & \textit{82.0} & \textit{71.2} & \textit{19.2} & \textit{81.2} & \textit{95.9} & \textit{78.2} & \textit{69.9} & \textit{46.8} & \textit{13.6} & \textit{67.0} \\
        \hline
        \hline
        Model & swg & tat & tuk & tzl & uig & uzb & war & wuu & xho & yid & yue & zsm & \multicolumn{2}{|c}{Avg.} \\
        \hline
        LASER & 46.0 & \textbf{31.1} & 20.7 & 44.7 & 45.2 & 18.7 & 13.6 & \textbf{87.7} & 8.5 & 5.7 & \textbf{90.0} & 96.4 & \multicolumn{2}{|c}{\textbf{47.5}} \\
        LASER2 & 29.9 & 20.2 & 14.8 & 40.9 & 38.2 & 15.7 & 5.4 & 52.4 & 3.9 & 3.4 & 64.9 & 89.0 & \multicolumn{2}{|c}{38.3} \\
        SBERT-distill & 33.9 & 17.8 & 24.1 & 41.3 & \textbf{65.5} & \textbf{32.6} & 11.4 & 82.7 & \textbf{11.6} & \textbf{52.7} & 84.4 & 95.6 & \multicolumn{2}{|c}{44.9} \\
        LaBSE-EMS-vanilla & 38.8 & 15.2 & 25.1 & 48.6 & 5.8 & 14.0 & 16.1 & 62.6 & 9.2 & 5.5 & 60.4 & 94.1 & \multicolumn{2}{|c}{34.5} \\
        LaBSE-EMS-MLP & 47.3 & 16.2 & 26.1 & 52.4 & 4.3 & 17.8 & 19.3 & 72.6 & 7.4 & 4.4 & 67.5 & 96.1 & \multicolumn{2}{|c}{38.0} \\
        LaBSE-EMS-joint (ours) & 56.7 & 22.4 & 29.8 & 59.1 & 6.8 & 24.1 & 27.5 & 82.1 & 9.5 & 9.6 & 76.9 & 96.7 & \multicolumn{2}{|c}{45.0} \\
        EMS (ours) & \textbf{58.9} & 25.7 & \textbf{30.5} & \textbf{61.1} & 8.1 & 23.7 & \textbf{28.7} & 82.9 & 8.5 & 11.6 & 78.9 & \textbf{97.0} & \multicolumn{2}{|c}{47.1} \\
        \hline
        \textit{LaBSE} & \textit{65.2} & \textit{87.9} & \textit{80.0} & \textit{63.0} & \textit{93.7} & \textit{86.8} & \textit{65.3} & \textit{90.3} & \textit{91.9} & \textit{91.0} & \textit{92.1} & \textit{96.9} & \multicolumn{2}{|c}{\textit{70.2}} \\
        \hline
    \end{tabular}
    }
\end{table*}

\subsection{Tatoeba Similarity Search}
\label{sec:5.1}
We use Tatoeba benchmark~\cite{DBLP:journals/tacl/ArtetxeS19} to evaluate the cross-lingual alignment between English and other 58 languages.\footnote{my, yo, and gu are not included in the Tatoeba benchmark.} Specifically, given a sentence in language $l_1$, we retrieve its translation from several sentences in language $l_2$. We use cosine similarity for retrieving sentences and report the average P@1 of $l_1\rightarrow l_2$ and $l_2\rightarrow l_1$ because both directions show similar precision considering a language pair.

As shown in Table~\ref{tatoeba}, in most languages, EMS achieves better retrieval precision than LASER, LASER2, SBERT-distill, LaBSE-EMS-vanilla, LaBSE-EMS-MLP, and performs comparably with LaBSE-EMS-joint. Observing the average score, 89.8, significantly outperforms LASER's 84.7, LASER2's 81.5, LaBSE-EMS-vanilla's 83.7, and is slightly higher than SBERT-distill and LaBSE-EMS-MLP. 

We further summarize the results of Table~\ref{tatoeba} in Table~\ref{tatoeba-all}. First, with regard to 15 main languages that mUSE~\cite{DBLP:conf/acl/YangCAGLCAYTSSK20} supports, our model achieves the best retrieval prevision, even better than LaBSE, which leveraged 6B training data and used a large batch size of 4,096 sentences. Second, with regard to 38 languages that XTREME~\cite{DBLP:conf/icml/HuRSNFJ20} supports, 48 languages that SBERT-distill supports, 43 languages for which we use less training data than LASER, and 42 languages for which all the models used training data over 300k, EMS consistently obtains higher retrieval precision than LASER, LASER2, SBERT-distill, LaBSE-EMS-vanilla, LaBSE-EMS-MLP, and performs on par with LaBSE-EMS-joint.

In addition, we observe similar results as compared with LaBSE for languages in which we used over 300k parallel sentences. This highlights the proposed model's efficiency in terms of data usage and computational resources for middle- and high-resource languages. Finally, with regard to 11 low-resource languages for which less than 300k training data are used in LASER, LASER2, LaBSE-EMS-vanilla, LaBSE-EMS-MLP, LaBSE-EMS-joint and EMS, EMS significantly outperforms than LASER, LASER2, LaBSE-EMS-vanilla, and LaBSE-EMS-MLP, whereas it is comparable with SBERT-distill and LaBSE-EMS-joint.\footnote{LASER, LASER2, LaBSE-EMS-x and EMS utilized less than 300k training data in the ``$<$300k(11)'' setting, whereas SBERT-distill and LaBSE significantly used more training data.}

Furthermore, we evaluate the other 54 unseen languages of our model (Table~\ref{unseen}). Although few languages are trained in LASER and LASER2, we observe that EMS and LaBSE-EMS-joint still yield results comparable with LASER and significantly better than LASER2 for these 54 languages. This indicates that EMS has cross-lingual transferability for unseen languages to a certain extent with the joint vocabulary.

In summary, the presented results on Tatoeba benchmarks highlight two key points: (1) EMS demonstrates superior data and computational efficiency, surpassing LASER, LASER2, and SBERT-distill; (2) AMS is not an optimal form of contrastive loss in a dual-encoder framework, while MLP layers, as introduced in our contrastive loss (Section~\ref{sec:3.3}), facilitate the effectiveness of AMS, showing that AMS is compatible with our framework under certain changes. More precisely with the second point, the inferior performance of LaBSE-EMS-vanilla demonstrates that LaBSE's AMS contrastive loss is dependent on LaBSE's extensive training data and large batch sizes. However, the incorporation of MLP layers (LaBSE-EMS-MLP) can enhance AMS's performance within EMS's efficient framework, and AMS's additive margin can enhance low-resource languages (see results of LaBSE-EMS-joint).

\begin{table*}[t]
    \centering
    \caption{\textbf{Average P@1 results of non-English language pairs on Flores benchmark.} \textbf{Bold} are the best precisions among all the models except LaBSE. ``mUSE'' and ``$<$300k'' respectively denote 182 high-resource and 240 low-resource language pairs. We additionally report the specific results of 8 randomly selected low-resource language pairs.}
    \label{flores}
    \resizebox{0.85\linewidth}{!}{
    \begin{tabular}{l|r|rrrrrrrr|r}
        \hline
        Model & mUSE (182) & af-gu & hi-hy & jv-ka & kk-mr & my-nb & sw-ta & te-tl & ur-yo & $<$300k (240) \\
        \hline
        LASER & 62.8 & 0.4 & 2.7 & 1.3 & 1.6 & 0.8 & 6.0 & 4.6 & 1.1 & 5.9 \\
        SBERT-distill & \textbf{99.7} & \textbf{84.7} & \textbf{99.1} & \textbf{46.6} & \textbf{82.5} & \textbf{97.3} & 14.6 & 23.2 & 18.3 & \textbf{60.6} \\
        LaBSE-EMS-joint (ours) & \textbf{99.7} & 13.6 & 89.6 & 39.3 & 72.7 & 3.5 & 32.7 & 38.0 & 20.1 & 46.6 \\
        EMS (ours) & \textbf{99.7} & 8.9 & 89.6 & 38.7 & 68.6 & 6.5 & \textbf{39.0} & \textbf{39.2} & \textbf{20.6} & 47.0 \\
        \hline
        \textit{LaBSE} & \textit{99.1} & \textit{100.0} & \textit{100.0} & \textit{99.9} & \textit{99.7} & \textit{99.5} & \textit{100.0} & \textit{99.9} & \textit{91.7} & \textit{98.9} \\
        \hline
    \end{tabular}
    }
\end{table*}

\subsection{Flores Similarity Search}
In this section, we assess the model's ability to perform cross-lingual retrieval for non-English language pairs using the Flores multilingual benchmark~\cite{DBLP:journals/tacl/GoyalGCCWJKRGF22}. Flores is an N-way evaluation dataset that consists of 200 languages, with 1,012 sentences per language. We assess two distinct groups of languages: (1) 14 main languages, excluding English, which are supported by mUSE; (2) 16 low-resource languages, which are designated as ``$<$300k'' in Section~\ref{sec:5.1}. This results in a total of 182 high-resource and 240 low-resource language pairs. We compute the P@1 metric for each language pair, as described in Section~\ref{sec:5.1}.

Table~\ref{flores} showcases the results for two groups of languages previously mentioned, along with eight randomly selected low-resource language pairs. Our analysis indicates that for 182 main non-English language pairs, SBERT-distill, LaBSE-EMS-joint, EMS, and LaBSE achieve nearly 100\% precision, whereas LASER demonstrated a precision of 62.8. These results highlight that SBERT-distill, LaBSE-EMS-joint, EMS, and LaBSE are English-independent models. Secondly, for low-resource language non-English language pairs, LASER performed poorly in retrieving the sentences accurately, whereas SBERT-distill, LaBSE-EMS-joint, and EMS delivered relatively good results. This demonstrates that the training objective of SBERT-distill, EMS, and LaBSE is conducive to generating language-agnostic embeddings. In contrast, the LASER model's translation objective still falls short of eliminating English as a pivot language for cross-lingual retrieval.

\subsection{BUCC: Bi-text Mining}

\begin{table*}[t]
    \centering
    \caption{\textbf{Extracted parallel sentence examples from BUCC that are not included in the official gold labels.}}
    \label{bucc-eg}
    \resizebox{0.95\linewidth}{!}{
    \begin{tabular}{ll}
        \hline
        \textit{Example 1} \\
        en & The Declaration of Brussels (1874) stated that the ``honours and rights of the family...should be respected.'' \\
        zh & \begin{CJK}{UTF8}{gkai}布鲁塞尔宣言（1874年）表示，``家庭荣誉和权利…应当受到尊重。''\end{CJK} \\
        \hline
        \textit{Example 2} \\
        en & In 2004. the E.U. undertook a major eastward enlargement, admitting ten new member states (eight of which were former communist states). \\
        zh & \begin{CJK}{UTF8}{gkai}2004年欧盟进行一次大规模东扩，接纳10个新成员国（其中的8个是前共产主义国家）。\end{CJK} \\
        \hline
        \textit{Example 3} \\
        en & In March 2013, Ban Ki-moon had also recommended to the Council that women raped in war have access to abortion services. \\
        zh & \begin{CJK}{UTF8}{gkai}2013年3月，潘基文同样建议安理会保证在战争中被强奸的妇女能享有堕胎服务。\end{CJK} \\
        \hline
    \end{tabular}
    }
\end{table*}

\begin{table}[t]
    \centering
    \caption{\textbf{F1 Scores on the BUCC benchmark.} \textbf{Bold} fonts denote the best precisions among mUSE, LASER, SBERT-distill, LaBSE-bilingual, LaBSE-EMS-joint, and EMS.}
    \label{bucc}
    \resizebox{0.95\linewidth}{!}{
    \begin{tabular}{l | rrrr | r}
        \hline
        Model & en-de & en-fr & en-ru & en-zh & Avg. \\
        \hline
        mUSE & 88.5	& 86.3 & 89.1 & 86.9 & 87.7 \\
        LASER & \textbf{95.4} & \textbf{92.4} & \textbf{92.3} & 91.2 & \textbf{92.8} \\
        SBERT-distill & 90.8 & 87.1 & 88.6 & 87.8 & 88.6 \\
        LaBSE-bilingual & 92.6 & 90.0 & 90.1 & \textbf{92.5} & 91.3 \\
        LaBSE-EMS-joint (ours) & 93.7 & 90.4 & 91.1 & 90.6 & 91.5 \\
        EMS (ours) & 93.3 & 90.2 & 91.3 & 92.1 & 91.7 \\
        \hline
        \textit{LaBSE} & \textit{95.9} & \textit{92.5} & \textit{92.4} & \textit{93.0} & \textit{93.5} \\
        \hline
    \end{tabular}
    }
\end{table}

Moreover, we evaluate the model's cross-lingual retrieval performance on BUCC benchmark~\cite{DBLP:conf/acl-bucc/ZweigenbaumSR17,zweigenbaum2018overview} that contains the comparable corpora with the size of 150k$\sim$1.2M for four language pairs: English--German, English--French, English--Russian, and English--Chinese. This task measures the model's ability to extract parallel sentences from comparable corpora. Following LASER and SBERT-distill, we use the margin-based scoring function~\cite{DBLP:conf/acl/ArtetxeS19} for mining parallel sentences. As the BUCC dataset mixes a significant number of monolingual sentences, we report F1 as the evaluation metric for this task following~\cite{DBLP:conf/emnlp/ReimersG20,DBLP:conf/acl/FengYCA022}, which differs from the one employed for Tatoeba and Flores.

Results measured using F1 are listed in Table~\ref{bucc}.\footnote{We use the code from \url{https://github.com/UKPLab/sentence-transformers/blob/master/examples/applications/parallel-sentence-mining/bucc2018.py}} We observe that EMS exhibits significantly higher results than mUSE~\cite{DBLP:conf/acl/YangCAGLCAYTSSK20} and SBERT-distill, and comparable results with LaBSE-lingual and LaBSE-EMS-joint. However, compared with LASER and LaBSE, EMS exhibits slightly poor performance. Such performance deterioration is negligible because it can be attributed to incorrect gold labels within the BUCC dataset, which is also mentioned in~\cite{DBLP:conf/emnlp/ReimersG20}. For example, three extracted sentence pairs listed in Table~\ref{bucc-eg} are translation pairs, whereas they are not contained in the official gold labels.

\begin{table}[t]
    \centering
    \caption{Cross-lingual sentence retrieval results on ParaCrawl. We report P@1 scores of 2,000 source queries while searching among 200k sentences in the target language. The best performance results among LASER, SBERT-distill, LaBSE-EMS-joint, and EMS are in \textbf{bold} font.}
    \label{paracrawl}
    \resizebox{\linewidth}{!}{
    \begin{tabular}{l|rrrrrrrr}
        \hline
        Model & bg & cs & da & de & el & es & et & fi \\
        \hline
        LASER & 90.5 & \textbf{87.8} & \textbf{86.1} & 89.4 & \textbf{85.3} & 89.4 & 87.6 & \textbf{83.4} \\
        SBERT-distill & 83.3 & 73.6 & 78.6 & 81.4 & 72.2 & 82.5 & 75.1 & 73.7 \\
        LaBSE-EMS-joint (ours) & 89.8 & 84.3 & 84.3 & 89.3 & 79.2 & 90.0 & 86.3 & 81.7 \\
        EMS (ours) & \textbf{90.9} & 85.5 & 85.1 & \textbf{90.1} & 81.4 & \textbf{90.9} & \textbf{87.7} & \textbf{83.4} \\
        \hline
        \textit{LaBSE} & \textit{91.2} & \textit{87.8} & \textit{88.9} & \textit{90.4} & \textit{85.3} & \textit{89.8} & \textit{88.3} & \textit{82.8} \\
        \hline
        \hline
        Model & fr & hr & hu & it & lt & lv & nl & pl \\
        \hline
        LASER & 90.9 & 87.1 & 86.8 & 82.5 & 89.0 & \textbf{84.8} & 88.3 & \textbf{81.9} \\
        SBERT-distill & 85.4 & 76.6 & 74.0 & 69.9 & 83.4 & 75.7 & 83.7 & 71.7 \\
        LaBSE-EMS-joint (ours) & 90.8 & 85.7 & 82.6 & 82.6 & 87.9 & 82.5 & 89.1 & 79.2 \\
        EMS (ours) & \textbf{91.6} & \textbf{87.3} & \textbf{87.5} & \textbf{83.5} & \textbf{90.5} & 84.0 & \textbf{90.6} & 81.5 \\
        \hline
        \textit{LaBSE} & \textit{90.5} & \textit{89.1} & \textit{84.5} & \textit{85.1} & \textit{91.0} & \textit{86.2} & \textit{89.5} & \textit{84.2} \\
        \hline
        \hline
        Model & pt & ro & sk & sl & sv && \multicolumn{2}{|c}{Avg.} \\
        \hline
        LASER & 90.9 & 85.2 & 87.9 & 88.9 & 85.3 && \multicolumn{2}{|c}{87.1} \\
        SBERT-distill & 86.2 & 80.4 & 79.2 & 80.0 & 79.8 && \multicolumn{2}{|c}{78.4} \\
        LaBSE-EMS-joint (ours) & 90.3 & 86.0 & 86.9 & 87.7 & 86.0 && \multicolumn{2}{|c}{85.8} \\
        EMS (ours) & \textbf{91.5} & \textbf{87.1} & \textbf{88.2} & \textbf{89.0} & \textbf{86.3} & & \multicolumn{2}{|c}{\textbf{87.3}} \\
        \hline
        \textit{LaBSE} & \textit{90.9} & \textit{88.2} & \textit{88.2} & \textit{89.6} & \textit{84.7} && \multicolumn{2}{|c}{\textit{87.9}} \\
        \hline
    \end{tabular}
    }
\end{table}

\subsection{Cross-Lingual Sentence Retrieval}
The Tatoeba benchmark supports the cross-lingual retrieval evaluation based on small-scale (1,000 sentences for most language pairs) data, whereas the BUCC benchmark supports retrieval from large-scale data for four language pairs. Therefore, we conduct a cross-lingual sentence retrieval evaluation based on large-scale comparable data for 21 language pairs.\footnote{In this study, we have chosen not to evaluate our model using the UN benchmark~\cite{DBLP:conf/lrec/ZiemskiJP16} as~\cite{DBLP:conf/acl/FengYCA022}. This decision is based on the fact that a portion of the UN benchmark data has been incorporated into our model's training dataset, which could potentially bias the evaluation results.} Based on our previous study~\cite{DBLP:conf/acl/MaoGCJK20}, given 2,000 sentences in language $l_1$, we conduct the translation retrieval from 200k candidate sentences in language $l_2$. Unlike our previous study, we used parallel sentences from ParaCrawl v5.0\footnote{\url{https://opus.nlpl.eu/ParaCrawl-v5.php}}~\cite{DBLP:conf/acl/BanonCHHHEFKKKO20} for evaluation because the previously used Europarl corpus is included in the training data in this study. We calculate P@1 for each language pair using margin-based scoring~\cite{DBLP:conf/acl/ArtetxeS19}. 

As reported in Table~\ref{paracrawl}, EMS performs significantly better than SBERT-distill LaBSE-EMS-joint, and is comparable with LASER and LaBSE. The 21 languages evaluated herein are trained with more than 300k parallel sentences, for which we used approximately half of the LASER's training data and a tiny fraction of the LaBSE's training data. This suggests that our training architecture and objective are rather effective for languages where we used a certain number of parallel sentences. Furthermore, the superior performance relative to LaBSE-EMS-joint underscores the limitations in the effectiveness of the additive margin introduced by LaBSE's AMS loss, when handling large-scale cross-lingual retrieval tasks.

\begin{table*}[t]
    \centering
    \caption{\textbf{MLDoc benchmark results (zero-shot scenario).} We report the mean accuracy of 5 runs. Best performance results of LASER, SBERT-distill, LaBSE-EMS-joint, and EMS are in \textbf{bold} font.}
    \label{mldoc}
    \resizebox{0.90\linewidth}{!}{
    \begin{tabular}{l | rr rr rr rr rr rr rr | r}
        \hline
        \multirow{2}{*}{Model} & \multicolumn{2}{c}{en-de} & \multicolumn{2}{c}{en-es} & \multicolumn{2}{c}{en-fr} & \multicolumn{2}{c}{en-it} & \multicolumn{2}{c}{en-ja} & \multicolumn{2}{c}{en-ru} & \multicolumn{2}{c|}{en-zh} & \multirow{2}{*}{Avg.} \\
        & $\rightarrow$ & $\leftarrow$ & $\rightarrow$ & $\leftarrow$ & $\rightarrow$ & $\leftarrow$ & $\rightarrow$ & $\leftarrow$ & $\rightarrow$ & $\leftarrow$ & $\rightarrow$ & $\leftarrow$ & $\rightarrow$ & $\leftarrow$ & \\
        \hline
        LASER & 86.3 & 76.7 & 76.2 & 68.1 & 82.1 & 75.7 & 70.3 & 69.8 & \textbf{71.5} & 59.8 & 64.6 & 68.9 & \textbf{77.7} & 67.3 & 72.5 \\
        SBERT-distill & 78.5 & 78.7 & 72.7 & 73.3 & 79.7 & 79.6 & 64.4 & 73.0 & 65.7 & 72.0 & 64.2 & 72.7 & 60.3 & 70.2 & 71.8 \\
        LaBSE-EMS-joint (ours) & 86.1 & 81.0 & 78.1 & \textbf{76.4} & \textbf{83.9} & 80.5 & 71.1 & 70.1 & 64.3 & \textbf{76.1} & 67.3 & 76.0 & 65.8 & \textbf{73.6} & 75.0 \\
        EMS (ours) & \textbf{87.6} & \textbf{81.1} & \textbf{82.0} & 75.5 & 82.9 & \textbf{80.6} & 70.4 & \textbf{73.6} & 67.0 & 72.3 & \textbf{68.5} & \textbf{77.5} & 68.6 & 69.1 & \textbf{75.5} \\
        \hline
        \textit{LaBSE} & \textit{87.2} & \textit{82.8} & \textit{78.8} & \textit{78.2} & \textit{87.3} & \textit{83.6} & \textit{74.1} & \textit{74.8} & \textit{73.4} & \textit{78.8} & \textit{74.6} & \textit{79.0} & \textit{85.3} & \textit{80.0} & \textit{79.9} \\
        \hline
    \end{tabular}
    }
\end{table*}

\begin{table*}[t]
    \centering
    \caption{\textbf{Results of the cross-lingual sentiment classification of Amazon Review version-1.} We report the mean accuracy of 5 runs. The best performance results of LASER, SBERT-distill, LaBSE-EMS-joint, and EMS are in \textbf{bold} font.}
    \label{cls1}
    \resizebox{\linewidth}{!}{
    \begin{tabular}{l|rrrrrr|rrrrrr|rrrrrr|r}
        \hline
        \multirow{3}{*}{Model} & \multicolumn{6}{c|}{en-de} & \multicolumn{6}{c|}{en-fr} & \multicolumn{6}{c|}{en-ja} & \multirow{3}{*}{Avg.} \\
        & \multicolumn{2}{c}{books} & \multicolumn{2}{c}{dvd} & \multicolumn{2}{c|}{music} & \multicolumn{2}{c}{books} & \multicolumn{2}{c}{dvd} & \multicolumn{2}{c|}{music} & \multicolumn{2}{c}{books} & \multicolumn{2}{c}{dvd} & \multicolumn{2}{c|}{music} &
        \\ 
        & $\rightarrow$ & $\leftarrow$ & $\rightarrow$ & $\leftarrow$ & $\rightarrow$ & $\leftarrow$ & $\rightarrow$ & $\leftarrow$ & $\rightarrow$ & $\leftarrow$ & $\rightarrow$ & $\leftarrow$ & $\rightarrow$ & $\leftarrow$ & $\rightarrow$ & $\leftarrow$ & $\rightarrow$ & $\leftarrow$ & \\
        \hline
        LASER & 78.3 & 76.0 & 73.7 & 73.4 & 76.1 & 77.2 & 77.2 & 77.4 & 76.8 & 75.4 & \textbf{75.8} & 76.6 & 72.0 & 72.9 & 73.0 & 70.9 & 75.5 & 75.5 & 75.2 \\
        SBERT-distill & 78.2 & 81.2 & 73.9 & \textbf{77.1} & 74.1 & 80.1 & 78.9 & 80.6 & 77.8 & 79.4 & 70.6 & 78.8 & 74.5 & \textbf{81.9} & \textbf{76.5} & 78.2 & 78.2 & 78.6 & 77.7 \\
        LaBSE-EMS-joint (ours) & \textbf{82.7} & 82.8 & \textbf{79.1} & 73.2 & 77.4 & \textbf{82.8} & 81.8 & 84.0 & \textbf{82.1} & 78.5 & 74.9 & 80.6 & \textbf{76.1} & 78.8 & 75.6 & 75.2 & 77.5 & 80.3 & 79.1 \\
        EMS (ours) & 82.3 & \textbf{84.9} & 77.0 & 76.7 & \textbf{78.8} & 81.9 & 80.4 & \textbf{84.6} & 78.0 & \textbf{81.1} & 74.7 & \textbf{83.0} & 75.6 & 79.5 & 75.4 & \textbf{79.2} & \textbf{79.2} & \textbf{80.9} & \textbf{79.6} \\
        \hline
        \textit{LaBSE} & \textit{82.2} & \textit{79.9} & \textit{77.1} & \textit{77.2} & \textit{79.0} & \textit{80.0} & \textit{83.2} & \textit{82.3} & \textit{81.0} & \textit{80.1} & \textit{77.9} & \textit{80.3} & \textit{78.0} & \textit{80.7} & \textit{77.7} & \textit{77.1} & \textit{81.6} & \textit{79.0} & \textit{79.7} \\
        \hline
    \end{tabular}
    }
\end{table*}

\subsection{MLDoc: Multilingual Document Classification}
Subsequently, we evaluate the model performance based on the MLDoc classification task. MLDoc\footnote{\url{https://github.com/facebookresearch/MLDoc}} is a benchmark to evaluate cross-lingual sentence representations, which contain datasets for eight languages~\cite{DBLP:journals/jmlr/LewisYRL04}. Following~\cite{DBLP:journals/tacl/ArtetxeS19}, we conduct the evaluation in a zero-shot manner using 1,000 sentences in language $l_1$ for training, 1,000 sentences in language $l_1$ for validation, and 4,000 sentences in language $l_2$ for the test. Specifically, we train a multilayer perceptron classifier based on source language representations and test the classifier for the target language. 

We list the average results of 5 runs for 7 language pairs and 14 directions in Table~\ref{mldoc}. We observe significantly higher accuracies of EMS in most directions than those of LASER and SBERT-distill, and comparable results with LaBSE-EMS-joint. These results demonstrate the effectiveness of the proposed training method. Although LASER yields better performance for English$\rightarrow$Japanese and English$\rightarrow$Chinese, it performs much worse in the reverse directions. We further calculate the average accuracy discrepancy between two directions for each language pair. LASER shows 7.3, whereas SBERT-distill is 3.5 and EMS is 4.7. This indicates that LASER is highly sensitive to the specific cross-lingual transfer direction, whereas SBERT-distill and EMS are much more robust.

\begin{table*}[t]
    \centering
    \caption{\textbf{Results of the cross-lingual sentiment classification of Amazon Review version-2.} We report the mean accuracy of 5 runs. The best performance results of LASER, SBERT-distill, LaBSE-EMS-joint, and EMS are in \textbf{bold} font.}
    \label{cls2}
    \resizebox{0.72\linewidth}{!}{
    \begin{tabular}{l | rr rr rr rr rr | r}
        \hline
        \multirow{2}{*}{Model} & \multicolumn{2}{c}{en-de} & \multicolumn{2}{c}{en-es} & \multicolumn{2}{c}{en-fr} & \multicolumn{2}{c}{en-ja} & \multicolumn{2}{c|}{en-zh} & \multirow{2}{*}{Avg.} \\
        & $\rightarrow$ & $\leftarrow$ & $\rightarrow$ & $\leftarrow$ & $\rightarrow$ & $\leftarrow$ & $\rightarrow$ & $\leftarrow$ & $\rightarrow$ & $\leftarrow$ & \\
        \hline
        LASER & 84.4 & 81.6 & 85.2 & 81.4 & 85.3 & 81.4 & 77.9 & 78.4 & 77.6 & 76.8 & 81.0 \\
        SBERT-distill & 85.8 & 85.6 & 87.0 & 85.8 & 86.8 & 84.6 & \textbf{81.7} & 83.8 & \textbf{81.6} & 81.3 & \textbf{84.4} \\
        LaBSE-EMS-joint (ours) & \textbf{86.4} & 83.5 & 85.8 & \textbf{86.0} & 86.0 & 85.1 & 79.4 & 80.8 & 78.0 & 81.8 & 83.3 \\
        EMS (ours) & 85.7 & \textbf{85.8} & \textbf{87.4} & 84.9 & \textbf{87.1} & \textbf{86.3} & 79.0 & \textbf{84.1} & 78.5 & \textbf{82.2} & 84.1 \\
        \hline
        \textit{LaBSE} & \textit{87.0} & \textit{84.5} & \textit{87.1} & \textit{85.3} & \textit{88.0} & \textit{84.7} & \textit{83.4} & \textit{82.0} & \textit{80.7} & \textit{79.9} & \textit{84.3} \\
        \hline
    \end{tabular}
    }
\end{table*}

\subsection{CLS: Cross-Lingual Sentiment Classification}
Moreover, we gauge the quality of language-agnostic sentence representation based on the sentiment classification task. We use the two versions of the Amazon review datasets for evaluation to conduct the zero-shot cross-lingual classification. The version-1 dataset~\cite{DBLP:conf/acl/PrettenhoferS10} includes the data for English--German, English—French, and English--Japanese on ``books,'' ``dvd,'' and ``music'' domains for each language pair. For each language pair and domain, we use 2,000 sentences in language $l_1$ for training, 2,000 sentences in language $l_1$ for validation, and 2,000 sentences in language $l_2$ for testing. However, the version-2 dataset~\cite{DBLP:conf/emnlp/KeungLSS20} includes five language pairs, whereas different genres of the reviews are mixed. For each language pair, we use 2,000 sentences for training, 4,000 sentences for validation, and 4,000 sentences for the test. Same as on MLDoc, we train a multi-layer perceptron using the language-agnostic sentence representations in language $l_1$ and test the classifier for another language. 

As listed in Tables~\ref{cls1} and~\ref{cls2}, EMS significantly outperforms LASER and performs comparably to LaBSE on the two versions of the datasets, which proves the effectiveness of EMS. SBERT-distill achieves comparable results on the version-2 dataset, whereas its performance negligibly deteriorates on the version-1 dataset. This can be attributed to SBERT-distill's capability of clustering similar sentences (Section 4.1 in~\cite{DBLP:conf/emnlp/ReimersG20}). On the version-1 dataset, each genre of the reviews is evaluated; more similar sentences in each genre compared with version-2 lead to lower classification accuracy for version-1. Moreover, EMS marginally surpasses LaBSE-EMS-joint in two benchmarks, suggesting that the additive margin in LaBSE's AMS does not enhance EMS's contrastive loss in classification tasks, aligning with the results on MLDoc.


\subsection{Ablation Study and Training Efficiency}

\begin{table*}[t]
    \centering
    \caption{\textbf{Ablation study of each model component, the AMS objective of LaBSE, and the computation resource.} Best performances are in \textbf{bold} font. The training efficiency is measured in seconds per 1k steps, utilizing four V100 GPUs.}
    \label{ablation}
    \resizebox{0.97\linewidth}{!}{
    \begin{tabular}{l | rrr |rr| rr | r}
        \hline
        \multirow{2}{*}{Model} & \multicolumn{3}{c|}{Tatoeba} & \multicolumn{2}{c|}{Flores} & \multicolumn{2}{c|}{MLDoc} & \multirow{2}{*}{Sec./1k Steps} \\
        & Avg. (58) & $>$300k (43) & $<$300k (15) & mUSE (182) & $<$300k (240) & en$\rightarrow$ X Avg. & X$\rightarrow$ en Avg. \\
        \hline
        EMS & 89.8 & 95.4 & \textbf{73.7} & \textbf{99.7} & \textbf{47.0} & \textbf{75.3} & 75.7 & 732 \\
        \hline
        \ \ $-$\ \textit{langs tok} & 89.3 & 95.4 & 71.8 & \textbf{99.7} & 46.9 & 73.5 & 74.4 & 725 \\
        \ \ $-$\ $\mathcal{L}_{cntrs}$ & 84.3 & 93.9 & 56.6 & 99.5 & 33.3 & 71.0 & 72.1 & 725 \\
        \ \ $-$\ $\mathcal{L}_{XTR}$ & 85.5 & 92.9 & 64.5 & 97.1 & 33.5 & 68.8 & 69.1 & 696 \\
        \ \ $-$\ $L_{cntrs\_mlp}$ & 86.3 & 93.8 & 64.6 & 99.1 & 39.0 & 69.8 & 73.8 & 727 \\
        \textit{share $L_{emb}$ params} & 85.1 & 93.8 & 60.3 & 98.8 & 32.2 & 68.5 & 71.1 & 730 \\
        \textit{replace XTR with UGT} & 86.9 & 94.5 & 64.9 & 99.5 & 40.6 & 71.1 & 73.8 & 731 \\
        \hline
        LaBSE-EMS-vanilla & 83.7 & 90.9 & 63.0 & 97.1 & 35.3 & 64.4 & 58.7 & 715 \\
        LaBSE-EMS-joint (ours) & \textbf{90.0} & \textbf{95.7} & \textbf{73.7} & \textbf{99.7} & 46.6 & 73.8 & \textbf{76.2} & 747 \\
        \hline
        \textit{replace V100 with A100} & 89.8 & 95.5 & 73.3 & \textbf{99.7} & 46.6 & 75.1 & 75.4 & - \\
        \hline
    \end{tabular}
    }
\end{table*}

\label{sec:5.7}
We conduct an ablation study to investigate the effectiveness of each model component and the computation resource. We report the results on the Tatoeba, Flores, and MLDoc benchmarks for cross-lingual sentence retrieval and classification tasks, respectively.

As listed in Table~\ref{ablation}, we observe that the performance significantly decreases on Tatoeba, Flores, and MLDoc benchmarks by removing the language token, sentence-level contrastive objective, XTR objective, or the MLP layers within the contrastive objective. Moreover, sharing the transformer embedding layer parameters with the $L_{emb}$ in the XTR objective and replacing XTR with UGT degrade the model performance. Among all these ablations, we observe a significant decrease in low-resource languages for training data less than 300k on both Tatoeba and Flores benchmarks, which indicates that the performance is more sensitive to model components on low-resource languages. This motivates future exploration to improve the performance of low-resource languages. 

By comparing ``$-$ $\mathcal{L}_{cntrs}$'' with ``$-$ $\mathcal{L}_{XTR}$,'' we observe superior performances of ``$-$ $\mathcal{L}_{cntrs}$'' on MLDoc and high-resource languages of Tatobea and Flores, and superior performances of ``$-$ $\mathcal{L}_{XTR}$'' on Tatoeba. This demonstrates that the generative objective contributes more to the classification of downstream tasks and the detection of parallel sentences high-resource language pairs, whereas the contrastive objective is more beneficial for the detection of parallel sentences of low-resource language pairs.

Moreover, we observe a negligible decrease in ``$-$ \textit{langs tok}'' on the Tatoeba and Flores benchmark. As the ground-truth label we designed for the XTR objective includes information on tokens in specific languages, the effect of the language token gradually diminishes during the model training. In our prior research, we recommended using UGT for bilingual settings; however, the current XTR in EMS surpasses it in three benchmarks. This may be because UGT is a more challenging task than XTR, which concurrently predicts the token distribution of the target language and a masked token, and the current model architecture may not be capable of accommodating numerous languages for UGT. Moreover, in our previous research~\cite{DBLP:conf/acl/MaoGCJK20}, we found that contrastive objectives negatively impacted classification tasks like MLDoc. However, in the current study with EMS in a massively multilingual context, both generative and contrastive objectives consistently enhance performance in retrieval and classification tasks. This improvement is likely due to the enhanced general capabilities of MSE, a result of exposure to massively multilingual data.

In terms of training efficiency, as detailed in Table~\ref{ablation}, the generative and contrastive objectives require extra 36 and 7 seconds per 1,000 steps, respectively. This is notably more efficient compared to using a transformer decoder for the generative objective in a translation task like LASER and LASER2, where the process could potentially double the training time.\footnote{A 6-layer transformer decoder demands more training time compared to a 6-layer transformer encoder due to its auto-regressive token generation process.} Our XTR approach enhances sentence embedding in a generative manner, bypassing the need for a transformer decoder in a dual-encoder setup. Incorporating an additive margin into the contrastive objective leads to diminished performance and efficiency in LaBSE-EMS-vanilla compared to $-$ $\mathcal{L}_{XTR}$, while LaBSE-EMS-joint shows only marginal improvements with a notable decrease in training efficiency relative to EMS. In addition, by replacing V100 GPUs with A100 and a larger batch size of 200 parallel sentences, only trivial performance fluctuation is observed, which suggests that EMS is robust to the computation resource.

\subsection{Case Study for the XTR Objective}
EMS utilizes an XTR objective that is independent of word order. To evaluate its robustness against sentences with identical word frequencies but differing semantics, we conducted a case study. This study compares EMS with LASER, SBERT-distill, and LaBSE using the following specific sentences designed for this purpose:

\noindent
(1). \textit{can you believe that what he actually did was steal the money she saved for the children?}

\noindent
(2). \textit{what can you actually believe she did was save the money for the children that he stole?}

We process the above sentences in their uncased form to guarantee an identical word bag for both sentences. Upon acquiring their sentence embeddings, we calculate the cosine similarity to determine the extent to which different MSE models perceived these two sentences as similar.

LASER, SBERT-distill, LaBSE, and EMS produce similarities of 0.90, 0.97, 0.95, and 0.97, respectively. This suggests that EMS struggles with such sentence pairs, despite the joint combined sentence-level contrastive objective should theoretically account for word order. Similarly, SBERT-distill and LaBSE, which do not incorporate word order-independent objectives like XTR, also fail to discern the semantic differences between the sentences. This indicates that sentence-level objectives within a dual-encoder architecture may not effectively address this issue. In contrast, LASER exhibits a lower similarity for this sentence pair, suggesting that its translation objective, which generates the target sentence word by word, might be more capable of resolving such issues. However, as these sentence pairs are relatively rare, further investigation into this limitation of the dual-encoder architecture is reserved for future research.

\section{Conclusion}
This study presented EMS, an efficient and effective method for MSE learning. To improve training efficiency in terms of data and computation while retaining the quality of MSE, we proposed a novel framework to train ``XTR'' generative and sentence-level contrastive objectives jointly. The empirical results based on four cross-lingual sentence retrieval tasks and three cross-lingual sentence classification tasks demonstrated the effectiveness of EMS. In future research, we aim to leverage LLMs for model initialization to refine sentence embeddings further and assess the benefits of a shallow decoder architecture~\cite{DBLP:conf/iclr/Kasai0PCS21} for MSE training. Additionally, we plan to streamline the model architecture through knowledge distillation, aiming for a more rapid inference experience.

\bibliographystyle{IEEEtran}
\bibliography{references}

\begin{IEEEbiography}[{\includegraphics[width=1in,height=1.25in,clip,keepaspectratio]{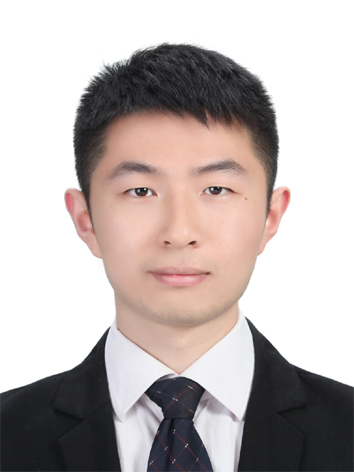}}]{Zhuoyuan Mao}
received the B.S. degree in Mathematics from East China University of Science and Technology, Shanghai, China, in 2017 and the M.S. and Ph.D. degrees in Informatics from Kyoto University, Kyoto, Japan, in 2021 and 2024, respectively. His research interests include natural language processing, machine translation, sentence representation learning, and multimodal machine learning.
\end{IEEEbiography}

\begin{IEEEbiography}[{\includegraphics[width=1in,height=1.25in,clip,keepaspectratio]{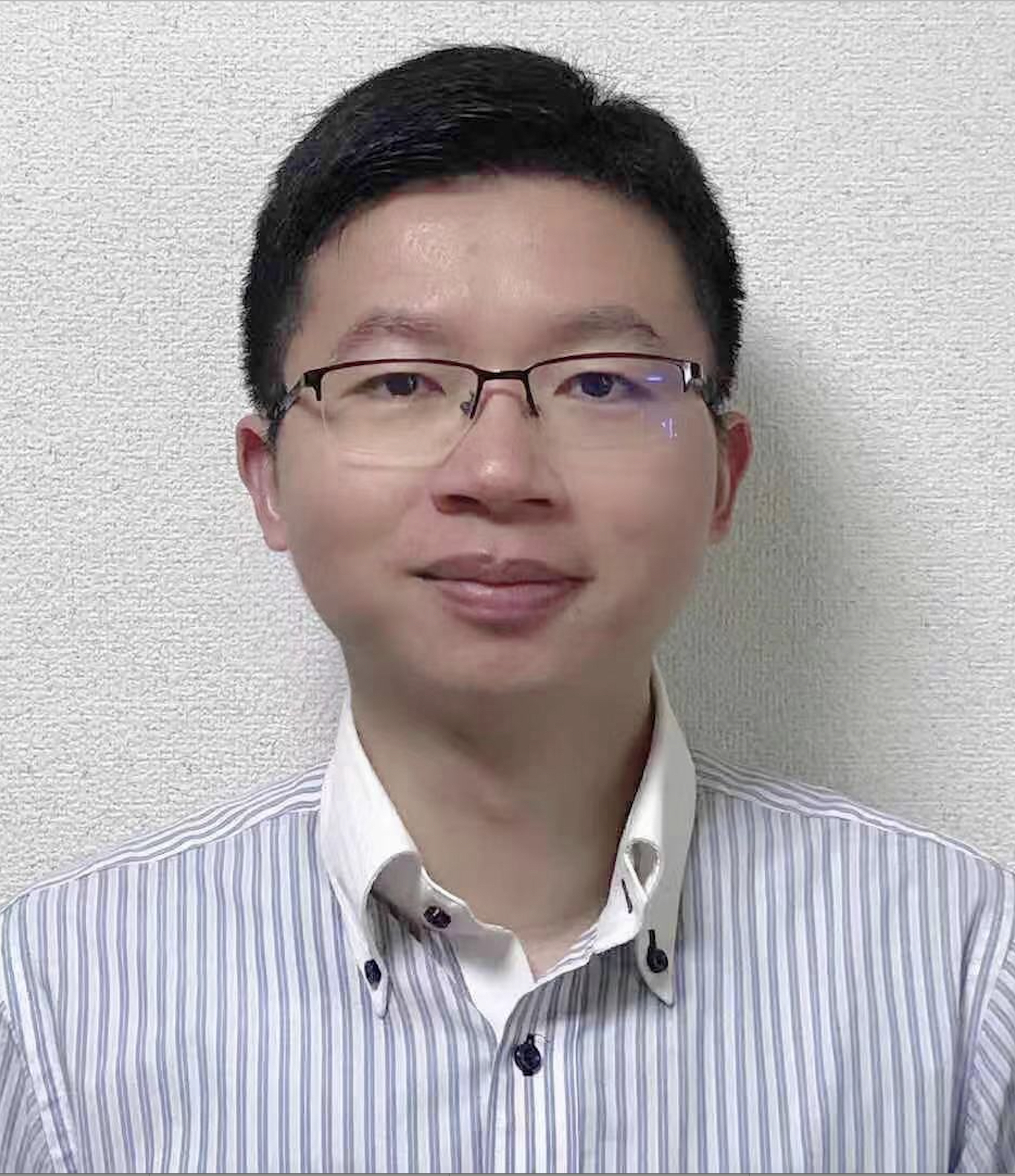}}]{Chenhui Chu}
received the B.S. degree in Software Engineering from Chongqing University, Chongqing, China, in 2008, and the M.S. and Ph.D. degrees in Informatics from Kyoto University, Kyoto, Japan, in 2012 and 2015, respectively. Currently, he is a program-specific associate professor at Kyoto University. His research interests include natural language processing, particularly machine translation and multimodal machine learning. 
\end{IEEEbiography}

\begin{IEEEbiography}[{\includegraphics[width=1in,height=1.25in,clip,keepaspectratio]{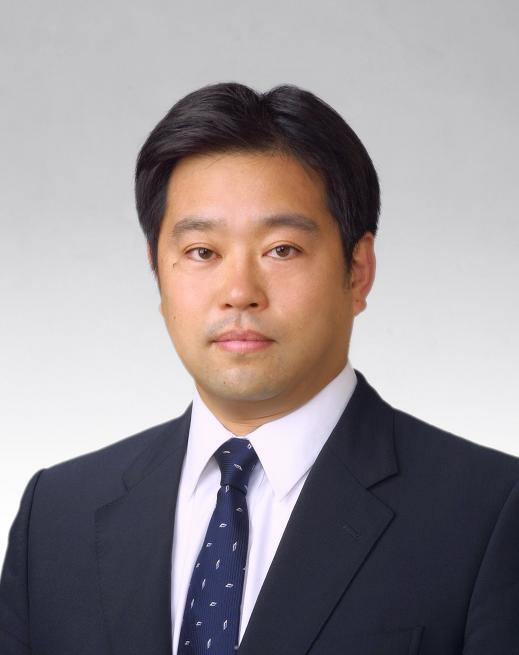}}]{Sadao Kurohashi}
received the B.S., M.S., and Ph.D. degrees in Electrical Engineering from Kyoto University, Kyoto, Japan, in 1989, 1991, and 1994, respectively. In 1994, he joined IRCS, University of Pennsylvania, PA, USA, as a visiting researcher. Currently, he is a professor at the Graduate School of Informatics at Kyoto University. His research interests include natural language processing, knowledge acquisition/representation, and information retrieval. He received the 10th anniversary best paper award from the Journal of Natural Language Processing in 2004, the Funai IT promotion award in 2009, and the IBM faculty award in 2009.
\end{IEEEbiography}

\vfill

\end{document}